%% file: bmvc_final.tex
\documentclass{bmvc2k}

\usepackage{graphicx}
\usepackage{multirow} 
\usepackage{subcaption}
\usepackage{algorithm}
\usepackage{algpseudocode}
\usepackage{afterpage}
\usepackage{textgreek}
\usepackage{booktabs}
\usepackage{amsthm}
\usepackage{enumitem} 
\usepackage{verbatim}
\usepackage{amssymb}
\usepackage{subfigure}  
\usepackage{wrapfig}
\newcommand{\tc}[1]{#1}

\newcommand{\xlt}[1]{#1}

\newcommand{\good}[1]{#1}

\makeatletter
\DeclareRobustCommand\onedot{\futurelet\@let@token\@onedot}
\def\@onedot{\ifx\@let@token.\else.\null\fi\xspace}

\def\eg{\emph{e.g}\onedot} 
\def\ie{\emph{i.e}\onedot}

\def\etc{\emph{etc}\onedot}

\newcommand{\reffig}[1]{Fig.~\ref{#1}}

\newcommand{\refsec}[1]{Section~\ref{#1}}

\def\sytheticGenerationName{\emph{Mesh2OB}}

\def\interactiveMethodName{\emph{MS\textsuperscript{3}PE}}
\def\deepModelName{\emph{TPE-Net}}

\def\interactionMechanism{\emph{MSIM}}

\def\sytheticBenchmarkName{\emph{OB-FUTURE}}

\def\realBenchmarkName{\emph{OB-LIGM}}

\def\subDIODE{\emph{OB-DIODE}}
\def\subEntitySeg{\emph{OB-EntitySeg}}

\newcommand{\corrauthor}{\hspace*{-1.8em}$^{\dagger}$ Corresponding author}
\begingroup
\footnotetext{\corrauthor}%
\addtocounter{footnote}{0} %
\endgroup

\title{Interactive Occlusion Boundary Estimation through Exploitation of Synthetic Data}

\addauthor{Lintao XU}{lintao.xu2@edu.univ-eiffel.fr}{1}
\addauthor{Chaohui WANG$^{\dagger}$}{chaohui.wang@univ-eiffel.fr}{1}

\addinstitution{
 LIGM, Univ Gustave Eiffel, \\
 École des Ponts, CNRS, \\ 
 Marne-la-Vallée, France
}

\runninghead{XU \& WANG}{IOBE through Exploitation of Synthetic Data}

\def\eg{\emph{e.g}\bmvaOneDot}

\begin{document}

\maketitle
\input{sec/0_abstract}

\input{sec/1_introduction}

\input{sec/2_related_work}
\input{sec/3_IOBE}

\input{sec/4_experiments}

\input{sec/5_conclusion}


\clearpage

\bibliography{egbib}
\end{document}

%% file: sec/0_abstract.tex
\begin{abstract}

Occlusion boundaries (OBs) geometrically localize occlusion events in 2D images and provide critical cues for scene understanding. In this paper, we present the first systematic study of \emph{Interactive Occlusion Boundary Estimation (IOBE)}, introducing~\textbf{\interactiveMethodName}~— a novel multi-scribble-guided deep-learning framework that advances IOBE through two key innovations: (1) an intuitive \textbf{m}ulti-\textbf{s}cribble interaction mechanism, and (2) a \textbf{3}-encoding-\textbf{p}ath network \textbf{e}nhanced with multi-scale strip convolutions. Our \textbf{\interactiveMethodName~}surpasses adapted baselines from seven state-of-the-art interactive segmentation methods, and demonstrates strong potential for OB benchmark construction through our real-user experiment.
Besides, to address the scarcity of well-annotated real-world data, we propose using synthetic data for training IOBE models, and developed\textbf{~\sytheticGenerationName}, the first automated tool for generating precise ground-truth OBs from 3D scenes with self-occlusions explicitly handled, enabling creation of the~\textbf{\sytheticBenchmarkName~}synthetic benchmark that facilitates generalizable training without domain adaptation. Finally, we introduce \textbf{\realBenchmarkName~}—  a high-quality real-world benchmark comprising 120 meticulously annotated high-resolution images advancing evaluation standards in OB research.
Source code and resources are available at \xlt{\url{https://github.com/xul-ops/IOBE}}.


\end{abstract}

%% file: sec/1_introduction.tex
\section{Introduction}
\label{sec:intro}

\begin{wrapfigure}[11]{r}{0.5\textwidth}
    \centering
    \vspace*{-5.5em}
    \includegraphics[width=0.40\textwidth]{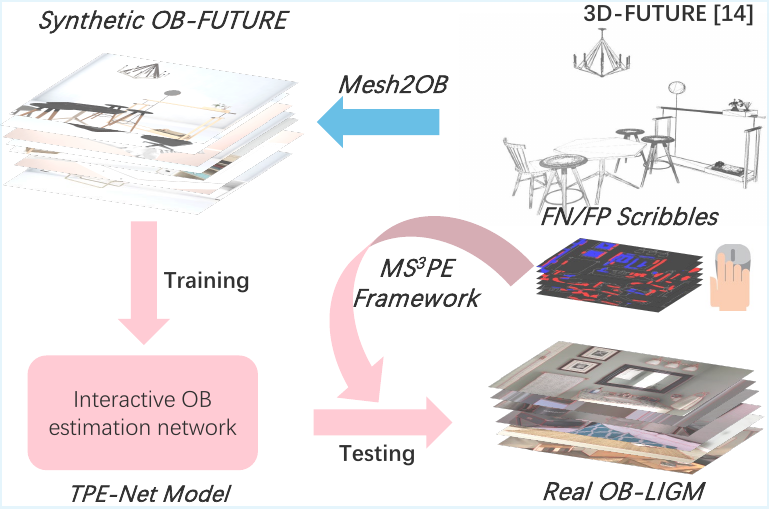}
    
    \captionsetup{format=plain,justification=justified,width=0.5\textwidth}
    \vspace{5pt}
    \caption{\footnotesize \textbf{Overall workflow.} During testing, given an image from the real-world benchmark~\realBenchmarkName,~\interactiveMethodName~(i) predicts initial occlusion boundaries (OBs); (ii) receives all false negative (FN) \& false positive (FP) scribbles from a human annotator, and (iii) outputs the refined result. For training, we generate synthetic data by applying~\sytheticGenerationName~to the \emph{3D-FUTURE} dataset~\cite{fu20213d}, creating the synthetic benchmark~\sytheticBenchmarkName.}

    \label{fig.teaser_iobe}
\end{wrapfigure}

\begin{figure}[t]
    \centering
    \small
    \resizebox{0.99\textwidth}{!}{
        \begin{tabular}{c}
            
            \hspace{-1.5em} (a)~\emph{CMU}~\cite{stein2009occlusion}
            \qquad\quad (b)~\emph{BSDS ownership}~\cite{ren2006figure}
            \qquad (c)~\emph{PIOD}~\cite{wang2016doc}
            \qquad\quad (d) \emph{NYUv2-OC++}~\cite{Ramamonjisoa_2020_CVPR}
            \qquad (e) \emph{iBims1\_OR}~\cite{qiu2020pixel}  
            \qquad (f) \emph{InteriorNet\_OR}~\cite{qiu2020pixel}
            \qquad (g) \sytheticBenchmarkName
            \qquad\quad (h) \realBenchmarkName
            
            \vspace{2pt}\\
            \includegraphics[width=0.25\linewidth]{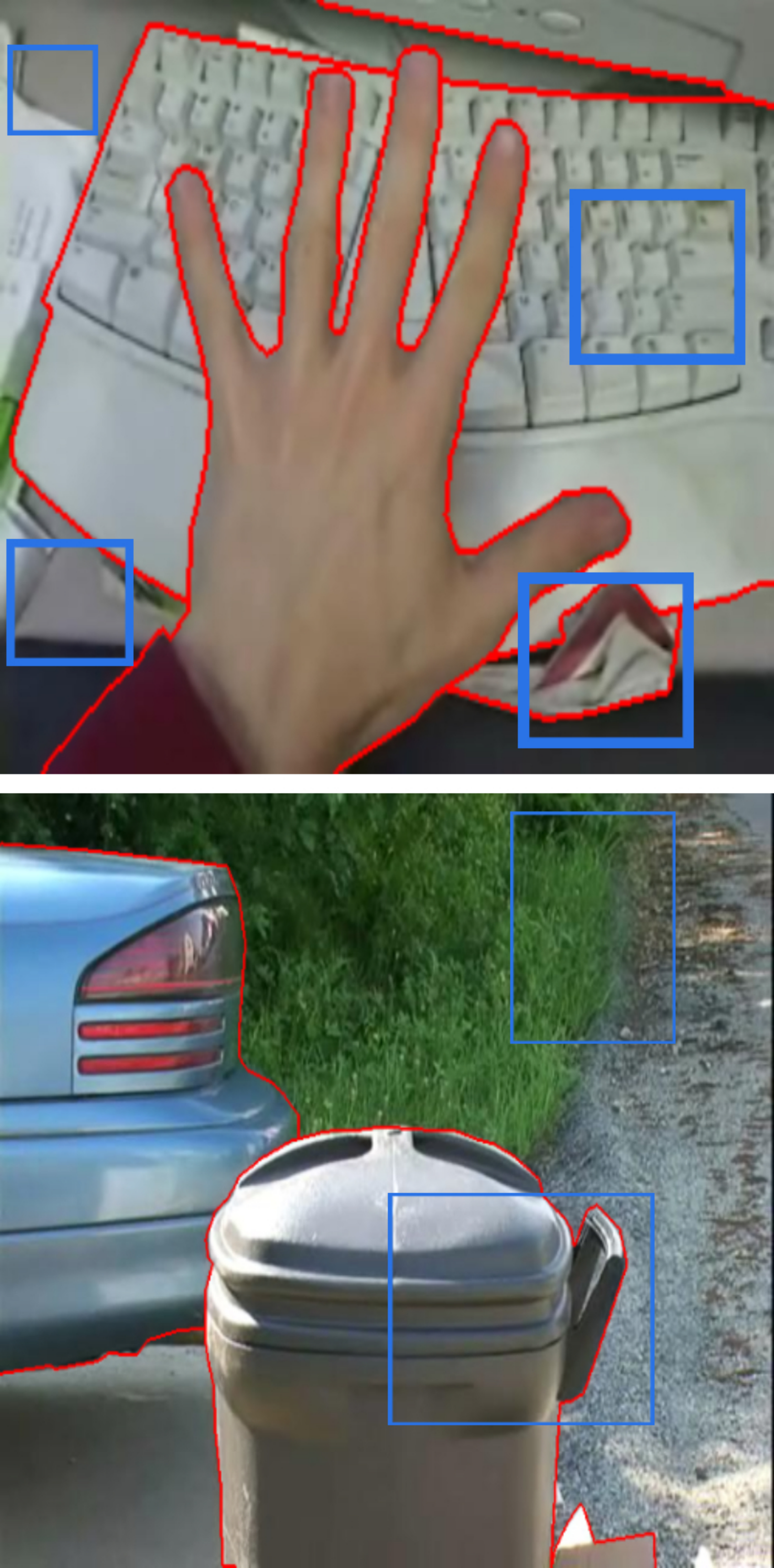}
            \includegraphics[width=0.25\linewidth]{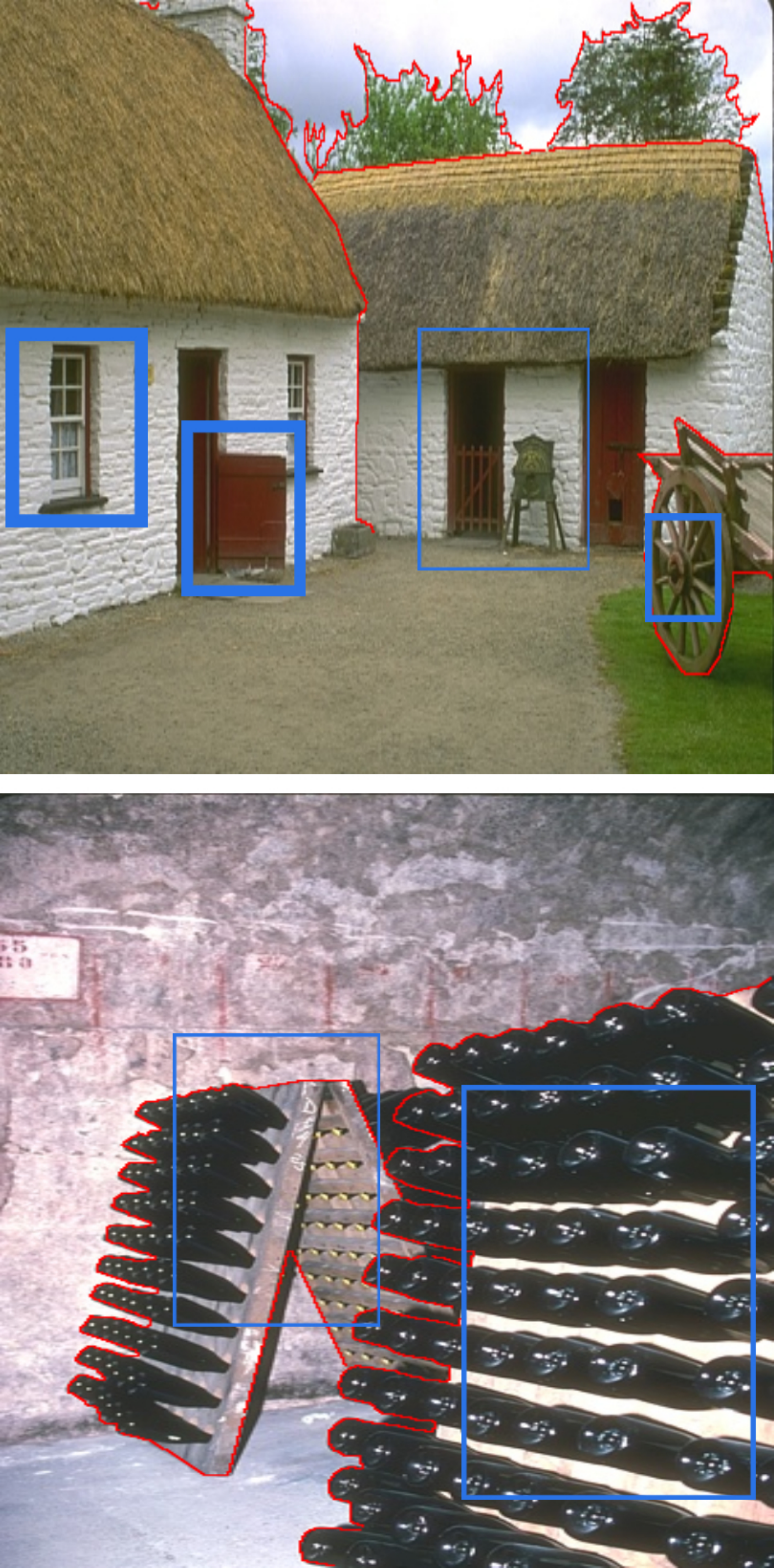}
            \includegraphics[width=0.25\linewidth]{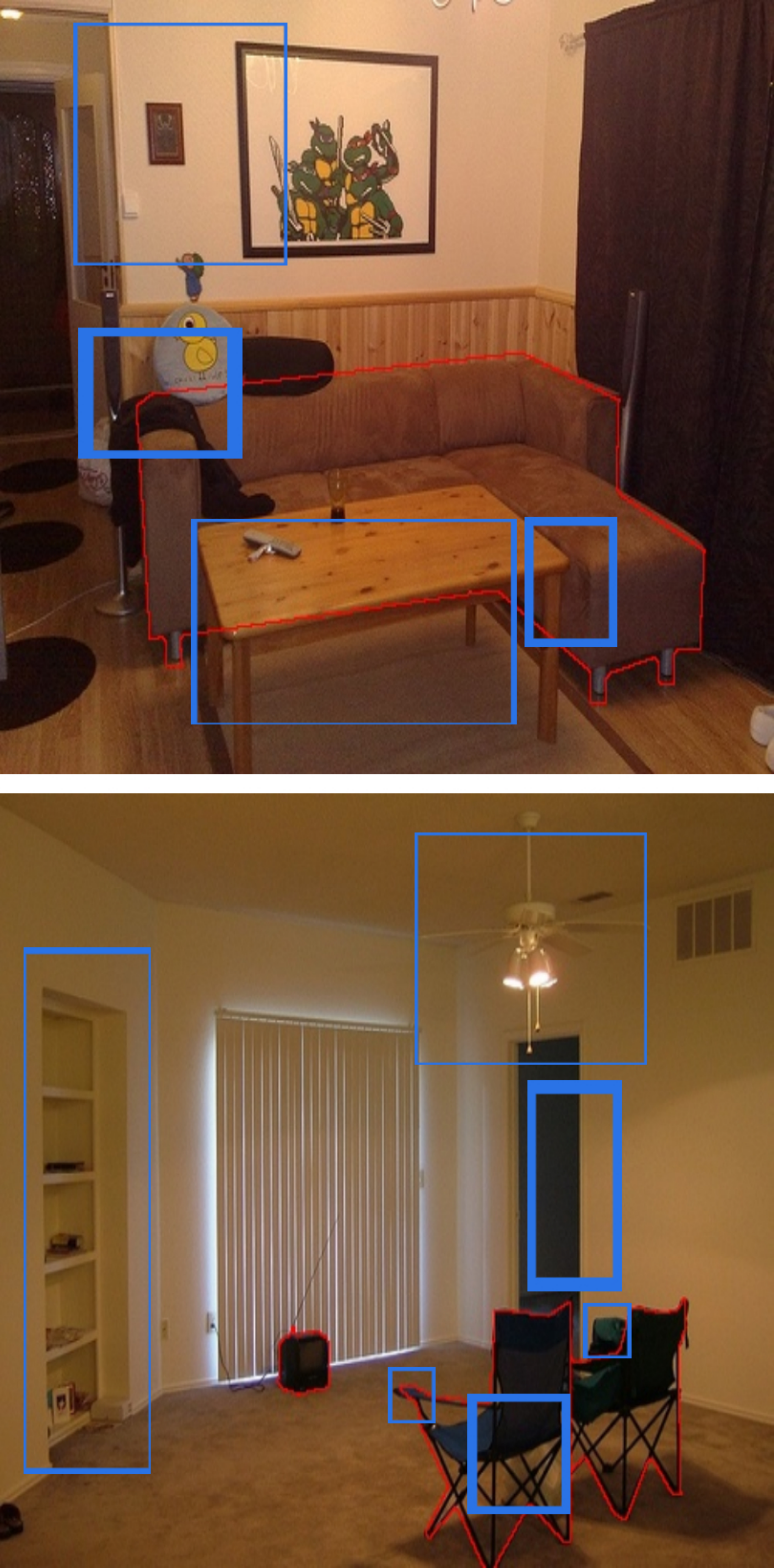}
            \includegraphics[width=0.25\linewidth]{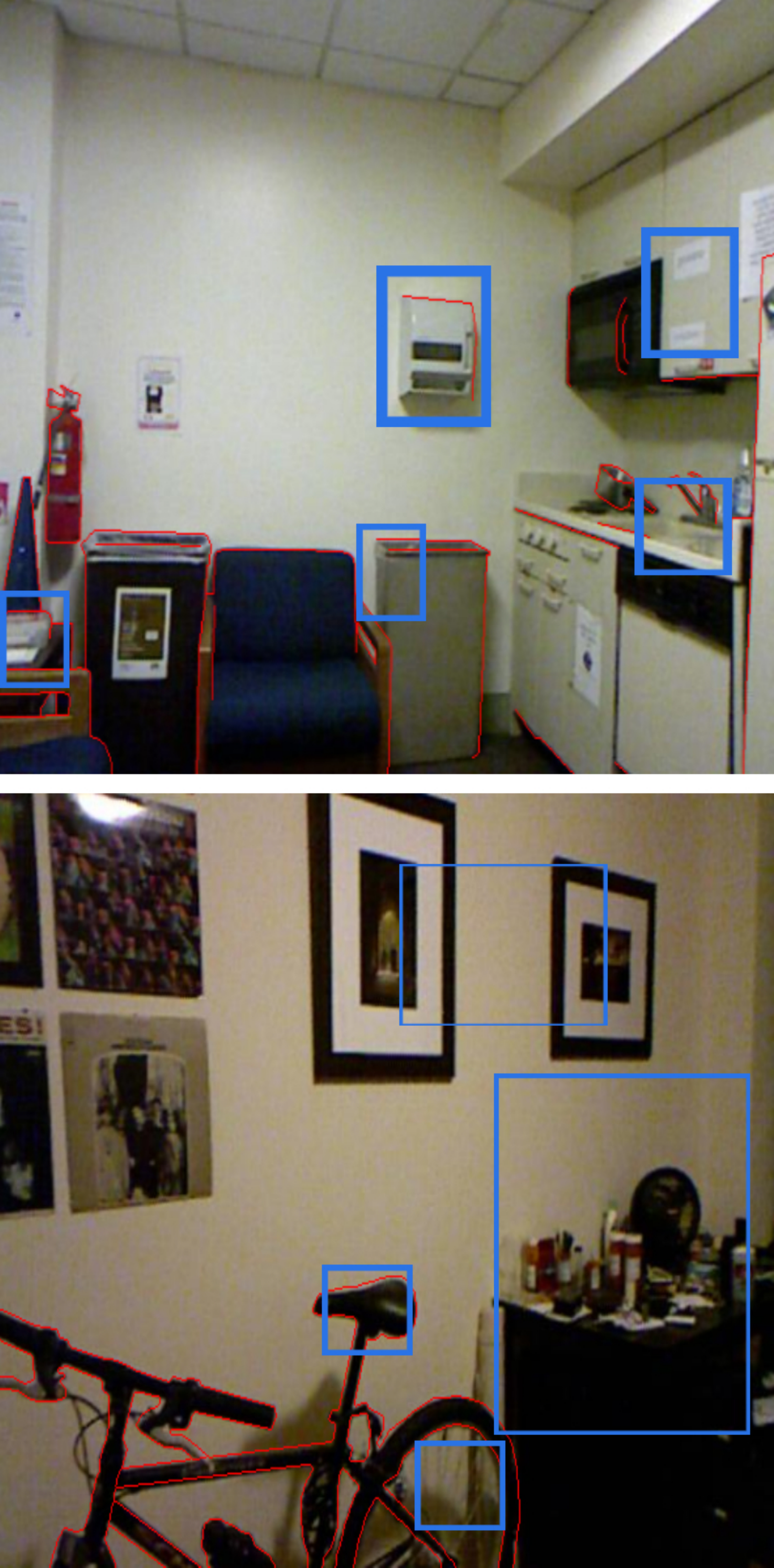}
            \includegraphics[width=0.25\linewidth]{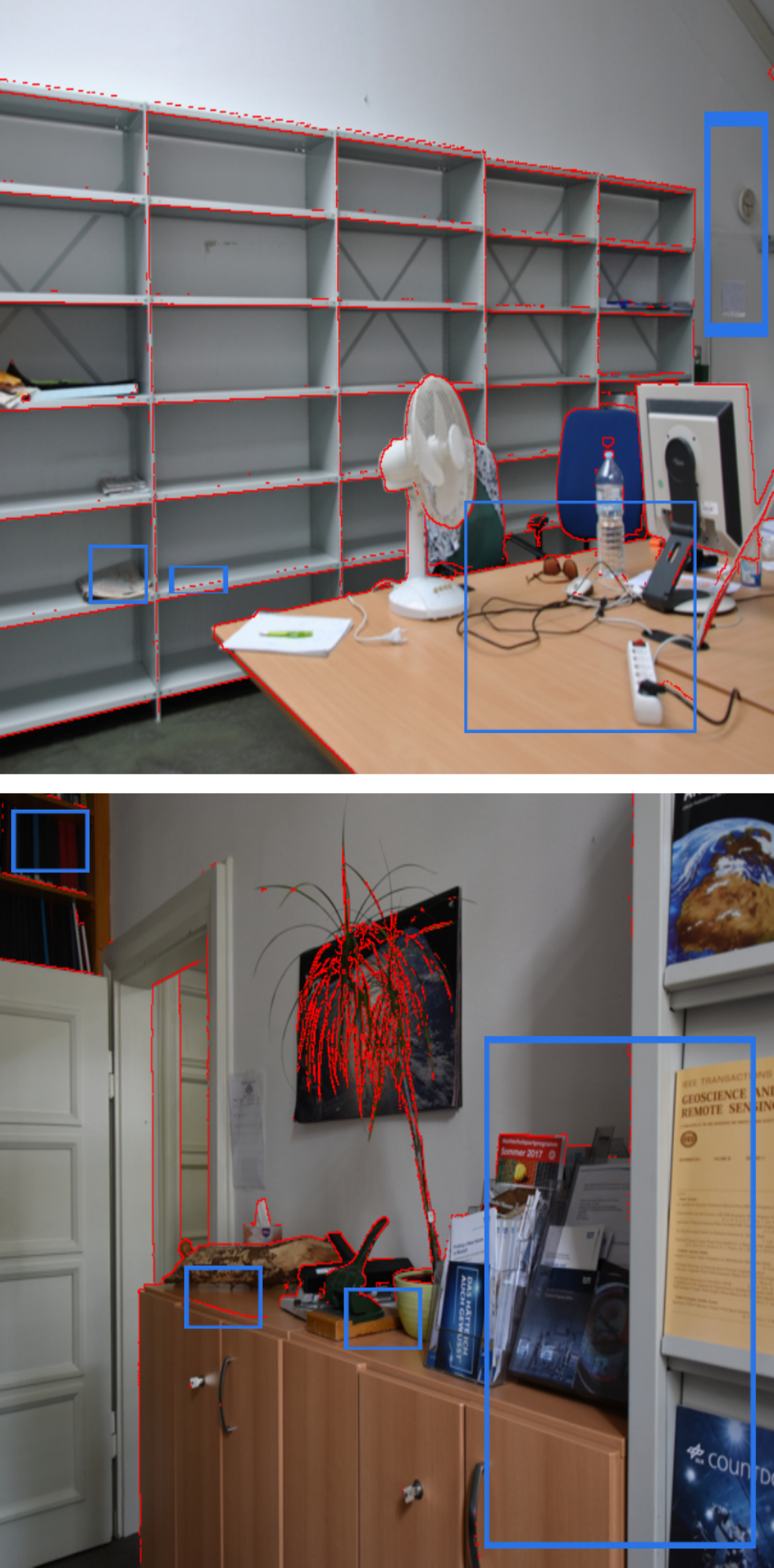}
            \includegraphics[width=0.25\linewidth]{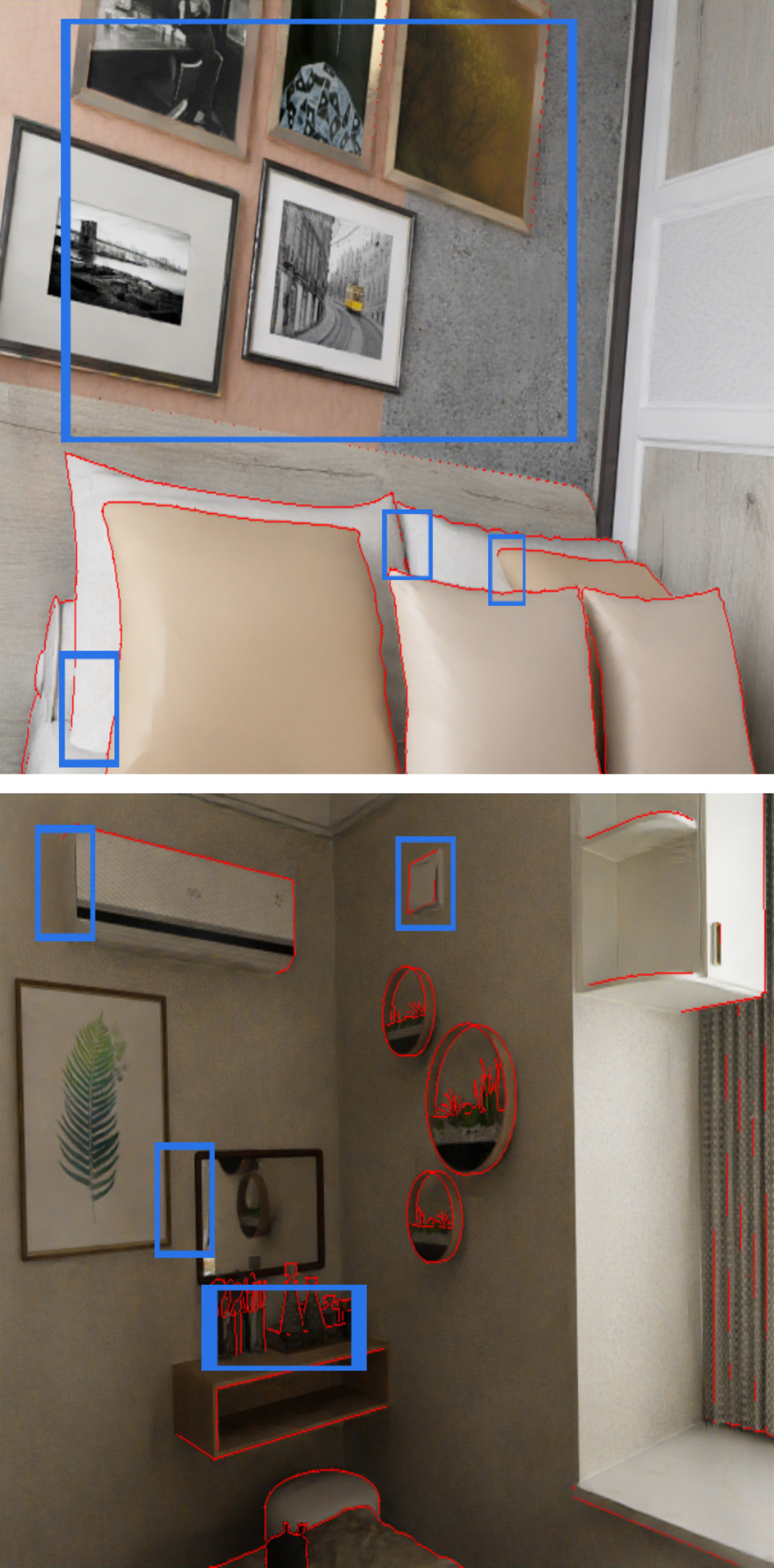}
            \includegraphics[width=0.25\linewidth]{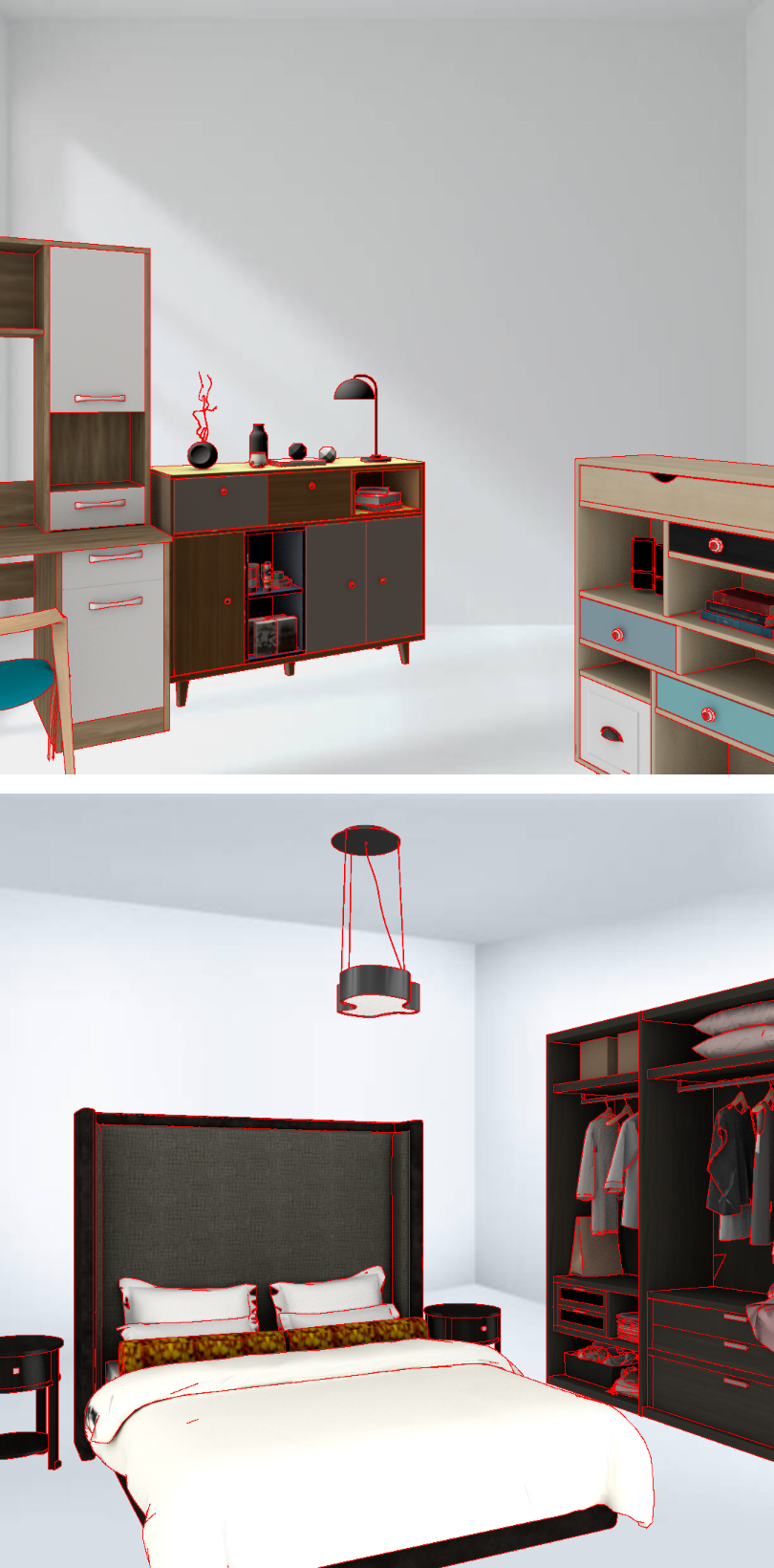}
            \includegraphics[width=0.25\linewidth]{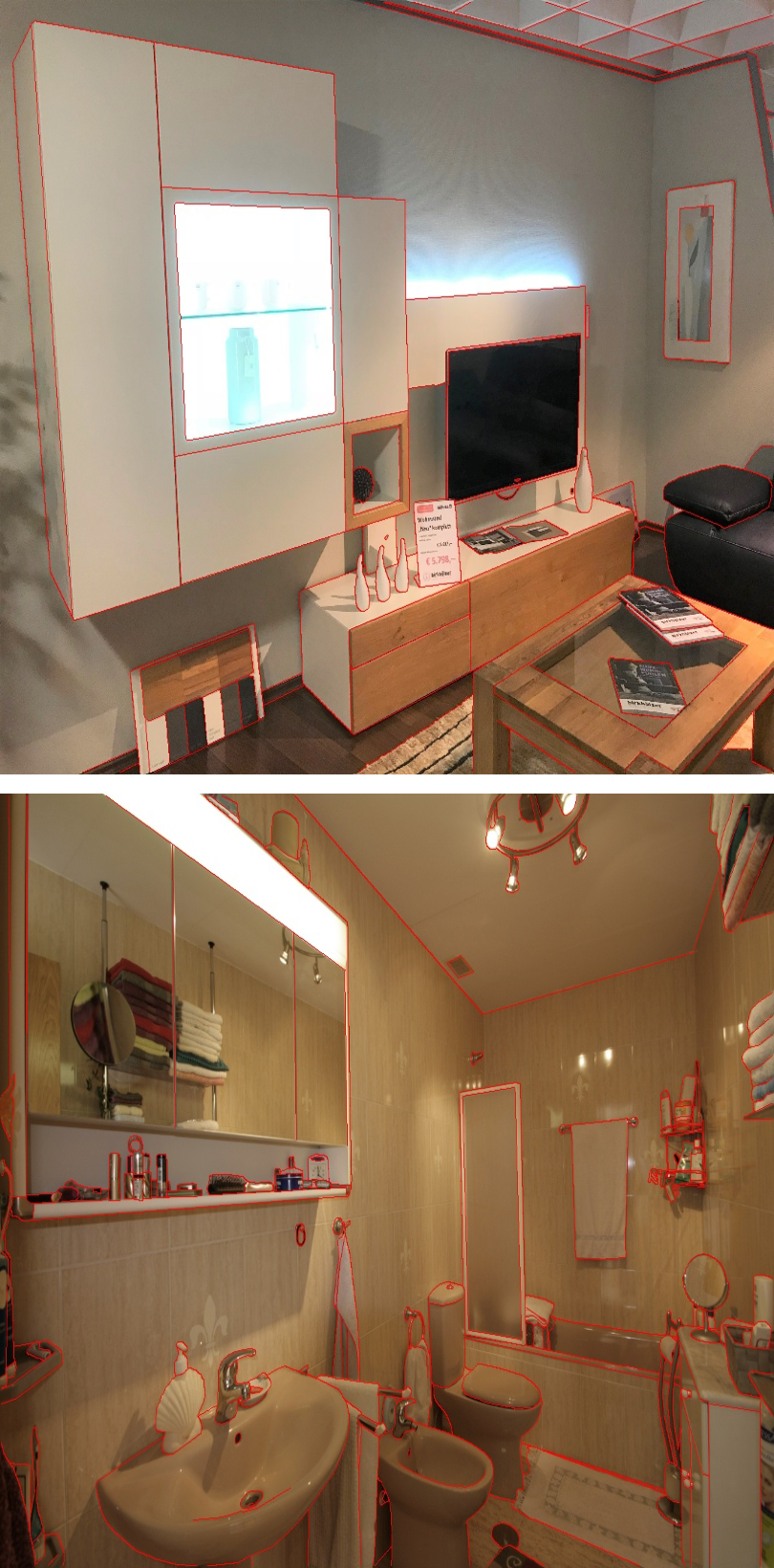}
            \\
        \vspace{1ex}
        \end{tabular}
    }
    \caption{\textbf{Benchmark visualization.} Red curves indicate OB ground truths. Blue boxes highlight representative regions containing missing, incomplete, and/or erroneous OBs.}
    \vspace{-14pt}
    \label{fig.occVis_compare}
\end{figure}


Occlusion is a common phenomenon in 2D images of natural scenes and poses a significant challenge to achieving high-quality visual understanding.
Despite the numerous works on occlusion handling~(\eg, \cite{qiu2020pixel, li2023neto,haouchine2015monocular,khirodkar2022occluded,ramamonjisoa2019sharpnet,guo2024redir,wu2025occfree_cvpr}), effectively addressing it remains a major challenge in computer vision.
As a key feature for characterizing occlusions, \emph{Occlusion Boundary} (OB) has been studied for a long time, both independently (\eg, \cite{black1992combining,apostoloff2005learning,sundberg2011occlusion,he2010occlusion,wang2020occlusion}) and in conjunction with orientations (\ie, occluder/occludee relationships) for OB (\eg, \cite{wang2016doc, wang2019doobnet, qiu2020pixel,feng2021mt,hambarde2024occlusion}). Various equivalents and analogues of it have been leveraged to improve the performance of scene understanding tasks, such as depth estimation, refinement, and ordering~\cite{xu2025modot,qiu2020pixel, hambarde2024occlusion, jia2012learning,ming2015monocular,zhu2017occlusion}, and 3D reconstruction and mesh recovery~\cite{castellani2002improving,keller20073d,Karsch_2013_CVPR,li2019high,yang2023sefd}.

A significant number of studies of OB estimation have been carried out (\eg, \cite{sundberg2011occlusion,he2010occlusion,hoiem2007recovering,wang2020occlusion,qiu2020pixel,feng2021mt,hambarde2024occlusion}), mainly based on the benchmarks shown in~\reffig{fig.occVis_compare}(a-d).
However, the Ground Truths (GTs) in those benchmarks were subjectively labeled without any unified pattern, making them open to debate.
For instance, lots of objects' OBs in~\emph{PIOD} are overlooked, and only visible depth discontinuities in~\emph{NYUv2-OC++} are annotated, which results in incomplete object contours.
The presence of subjective bias in their GTs provided, along with the insufficient quality/precision of the GTs (see Table~\ref{tab-db_compare}), hinders the scientific and systematic study of OB.
Later, a \emph{mathematical definition} has been introduced in~\cite{wang2020occlusion}, which specifies that a small section of an OB represents an occlusion event caused by the 3D surfaces corresponding to the 2D regions on either side of the boundary, and that all occlusion events in the entire image are considered together.
This definition establishes that OBs are derived from the 3D geometry of the scene relative to the camera. Notably, OBs can arise in both inter-object regions (\eg, between distinct objects) and intra-object regions, where \emph{self-occlusion boundaries} occur due to occlusions between constituent parts of a single object. This leads to a key property of OBs: their \emph{closure-indeterminacy}—they may appear in either open or closed contours in the 2D image plane (\reffig{fig.occVis_compare}~(g,h)).
In practice, a non-closed OB often partially or totally lies in a region with a single photometric model (\eg, the self-occlusion boundaries of the washbasin and the flush toilet in the bottom image of~\reffig{fig.occVis_compare}~(h)), as the two sides belong to the same object.

These properties fundamentally distinguish OB estimation from those segmentation tasks that partition an image into connected regions, making it sufficiently unique to warrant independent study as a distinct research problem.

Interactive Segmentation (IS) methods (\eg, \cite{sofiiuk2022reviving,liu2023simpleclick,Zhao2024graco}) aim to segment specific objects or parts in an image through minimal user input. As a fundamental task in computer vision, IS has found widespread applications in photo editing and medical image analysis. More importantly, it plays a critical role in efficiently building and expanding datasets, serving as a pixel-level interactive AI system that follows human intent. Interactive mechanisms in IS leverage human guidance to resolve ambiguities that automated models often struggle with, enabling the correction of prediction errors and the segmentation from unseen classes.
Like IS, OB estimation benefits from human-in-the-loop interaction to disambiguate challenging cases.
OB cues are typically subtle and can be easily confused with edges resulting from lighting variations, texture patterns, or reflections. Additionally, they may fail to capture self-occlusion boundaries that occur within object contours.
In addition, OB estimation suffers from a scarcity of high-quality annotated data. 
In such cases, user-provided interactions offer attention priors that help models focus on true occlusion cues grounded in the 3D structure of the scene. By incorporating interaction, our IOBE approach enables users to iteratively refine OB predictions, enabling the model to more effectively discover undetected OBs and differentiate OBs from other edge types. This not only improves prediction accuracy but also supports scalable and practical OB dataset annotation workflows.

The above considerations have motivated our study of \textbf{\emph{interactive OB estimation (IOBE)}}, aiming to improve OB estimation accuracy through human interaction while efficiently constructing OB benchmarks. To the best of our knowledge, this is the \textbf{pioneering} work on IOBE, and our main contributions are threefold:


\vspace{-6pt}
\begin{enumerate}

\item We predominantly propose\textbf{~\interactiveMethodName}, a multiple-scribble-guided deep-learning framework for IOBE, composed of (1) an intuitive \emph{\textbf{m}ulti-\textbf{s}cribble \textbf{i}nteraction \textbf{m}echanism} (\textbf{\interactionMechanism)}, and (2) a \textbf{t}hree-encoding-\textbf{p}ath network \textbf{e}nhanced with multi-scale strip convolutions \textbf{(\deepModelName)}.
Extensive experiments validate the effectiveness of the proposed~\deepModelName~and~\interactionMechanism, demonstrating that our method significantly outperforms adapted baselines from seven state-of-the-art interactive segmentation methods~\cite{lin2020fcanet,liew2021deepthin,sofiiuk2022reviving,chen2021conditional,chen2022focalclick,liu2023simpleclick,lin2025adapclick}, while showing strong potential for OB benchmark construction.

\vspace{-3pt}
\item 
To address the scarcity of well-annotated real-world data, we propose using synthetic data for training IOBE models, thanks to the particularity that OBs can be unambiguously derived from the 3D scene. 
To this end, building upon~\cite{wang2020occlusion}'s definition, we developed\textbf{~\sytheticGenerationName}, an efficient tool for automatically generating OBs from mesh-based 3D scene data with \textbf{self-occlusions} explicitly handled.
By applying~\sytheticGenerationName~on the \emph{3D-FUTURE}~\cite{fu20213d}, we established~\textbf{\sytheticBenchmarkName}, a benchmark consisting of 19,186 high-quality synthetic indoor scenes. Extensive experiments demonstrate its effectiveness as a generalizable training resource.
To the best of our knowledge, we are the first to generate the OBs for 2D images directly from 3D scene data where self-occlusions are considered. By effectively leveraging the 3D scene's geometric information, \sytheticGenerationName~achieves significantly higher quality than~\cite{qiu2020pixel}, the only previous method which handles self-occlusions from depth and normal maps.

\vspace{-3pt}
\item 

We introduce~\textbf{\realBenchmarkName}, a real-world benchmark comprising two subsets, \subEntitySeg\ and \subDIODE, 
constructed through meticulous manual annotation of GT OBs on 70 and 50 high-resolution indoor images from two existing datasets~\cite{vasiljevic2019diode, qi2022fine}  designed for different tasks.
\realBenchmarkName~significantly surpasses prior benchmarks in annotation quality, enabling a more compelling and robust evaluation in OB research.

\end{enumerate}
\vspace{-6pt}

%% file: sec/2_related_work.tex
\vspace{-11pt}
\section{Related Work}
\label{sec:related_work}
\vspace{-7pt}

\begin{table*}[htbp]
\vspace{-6pt}
  \centering
  \small
  \caption{\textbf{Comparison with previous OB benchmarks.} \footnotesize``\emph{syn}'' refers to synthetic. OB ground truth annotations in our proposed datasets are high-quality and densely labeled across high-resolution scenes, particularly capturing full-image intra-object self-occlusions.}
  \vspace{10pt}
  \label{tab-db_compare}
  \resizebox{0.95\textwidth}{!}
  {
  \begin{tabular} {lccccccc} 
      \toprule 
      \multicolumn{1}{l}{} & \multicolumn{4}{c}{\textbf{General Information}} & \multicolumn{3}{c}{\textbf{OB GT's Characteristics}} \\
      \cmidrule(lr){2-5} \cmidrule(lr){6-8}
      \multirow{2}{*}{\textbf{Datasets}} & \textbf{Data}  & \textbf{Image} & \textbf{Scene} & \textbf{GT}  & \textbf{Self-occlusion} & \textbf{Object-contour} & \textbf{Full-image}   \\
      & \textbf{Volume} & \textbf{Resolution} & \textbf{Type} & \textbf{Annotation} & \textbf{Boundary} & \textbf{Completeness} & \textbf{Annotation}  \\    
      \midrule
      \emph{CMU}~\cite{stein2009occlusion} & 30 & (640, 480) & real & manual  & ignored  & $\checkmark$  & $\mathbf{\times}$  \\
      \emph{BSDS Ownership}~\cite{ren2006figure} & 200 & (481, 321) & real & manual & ignored  & $\checkmark$ & $\checkmark$  \\
      \emph{PIOD}~\cite{wang2016doc} & 10,000 & (475, 390) & real & manual & ignored  & $\checkmark$ & $\mathbf{\times}$ \\
      \emph{NYUv2-OC++}~\cite{Ramamonjisoa_2020_CVPR} & 654 & (592, 440) & real & manual  & incomplete & $\mathbf{\times}$ & $\mathbf{\times}$  \\
      \emph{iBims1\_OR}~\cite{qiu2020pixel} & 100 & (640, 480) & real & syn & incomplete  & $\mathbf{\times}$ & $\mathbf{\times}$  \\
      \emph{InteriorNet\_OR}~\cite{qiu2020pixel} & 10,000 & (640, 480) & syn & syn & incomplete & $\mathbf{\times}$ & $\mathbf{\times}$  \\
      \midrule
      \textbf{our~\sytheticBenchmarkName} & 19,186 & (1080, 1080) & syn & syn & \textbf{complete} & $\checkmark$ & $\checkmark$ \\
      \textbf{our~\subDIODE} & 50 & (1024, 768) & real & manual & \textbf{complete}  & $\checkmark$ & $\checkmark$  \\
      \textbf{our~\subEntitySeg} & 70 & (982, 882) & real & manual & \textbf{complete} & $\checkmark$ & $\checkmark$ \\
      \bottomrule 
  \end{tabular}
  }
  \vspace{-6pt}
\end{table*}

\noindent\textbf{Fully-automatic OB Estimation} remains a persistent challenge in computer vision, with numerous previous works addressing this issue.
One subset of methods (\eg,~\cite{apostoloff2005learning,feldman2008motion,stein2009occlusion,sargin2009probabilistic,sundberg2011occlusion,he2010occlusion,jacobson2011online, wang2020occlusion}) estimates OB maps from image sequences. Another subset of approaches (\eg,~\cite{hoiem2007recovering, ren2006figure, raskar2004non, teo2015fast, wang2016doc,wang2019doobnet,lu2019occlusion,feng2021mt,qiu2020pixel,hambarde2024occlusion,Ramamonjisoa_2020_CVPR}) infers OB maps from 2D monocular images, either independently or along with the occluder/occludee relationships for OBs.
To name a few,~\cite{hoiem2007recovering} recovered
OBs using the traditional edge and region cues together with 3D surface and depth cues.
~\cite{teo2015fast} imposed a border ownership structure and simultaneously detected both boundaries and border ownership using structured random forests.
\cite{wang2019doobnet} designed a one-stream deep method and a novel attention loss to tackle extreme boundary/non-boundary class imbalance.
\cite{Ramamonjisoa_2020_CVPR} utilized the displacement field to refine depth thus obtaining more accurate OBs.
MTORL~\cite{feng2021mt} proposed \emph{OPNet} which only shares backbone features in the network and uses the side output to enhance the final OB prediction.

\vspace{3pt}
\noindent \textbf{OB benchmarks}\footnote{More details are included in the Supplementary Materials (referred to as \emph{Supp.} hereafter).} include \emph{CMU}~\cite{stein2009occlusion}, \emph{BSDS ownership}~\cite{ren2006figure}, \emph{PIOD}~\cite{wang2016doc}, \emph{NYUv2-OC++}~\cite{Ramamonjisoa_2020_CVPR} \emph{iBims1\_OR}~\cite{qiu2020pixel} and \emph{InteriorNet\_OR}~\cite{qiu2020pixel}. The final two benchmarks are constructed using source images from~\cite{koch2018evaluation,li2018interiornet}, with synthetic GT boundaries generated from depth and normal maps via the method of~\cite{qiu2020pixel}. In all these benchmarks, self-occlusion boundaries are either totally ignored or only partially considered. More differences are presented in Table~\ref{tab-db_compare}.

\vspace{3pt}
\noindent \textbf{Interactive Segmentation (IS)} aims to segment single objects (\eg, \cite{xu2016deep, sofiiuk2022reviving}) or multiple objects (\eg, \cite{li2022efficient, lee2022interactive}) by incorporating user interactions such as clicks (\eg, \cite{lin2020fcanet}), scribbles (\eg, \cite{chen2023scribbleseg}), and bounding boxes (\eg, \cite{zhang2020interactive}).
It has also garnered significant attention in medical image segmentation~(\eg, \cite{wang2018interactive, luo2021mideepseg,huang2024efficientmedical}).
To date, the topic has seen significant advancements, along with the development of numerous deep learning-based methods (\eg, \cite{jang2019interactive, chen2022focalclick, liu2023simpleclick, du2023efficient, zhou2023interactive, Myers-Dean_2024_WACV, Zhao2024graco,lee2024mfp,M2N22025}).
Scribble-based IS typically places scribbles inside and/or outside the object of interest, rather than on its boundaries. These scribbles act as corrective guides to refine prediction masks in error-prone areas \cite{Agustsson_2019_CVPR, andriluka2020efficient}, or as general indicators to help models learn the complete object segmentation mask \cite{galeev2023contour, chen2023scribbleseg}.
Unfortunately, these scribble-based techniques require that object contours be fully closed, making them unsuitable for our problem.
Several IS methods~\cite{le2018interactive, majumder2020two, dupont2021ucp, maninis2018deep, liew2021deepthin} apply clicks on object contours for better performance, requiring much greater precision than the scribbles used in our approach.

The extension of IS to the full-image scenario is more relevant to our problem, yet it has been much less explored. Early works, such as~\cite{nieuwenhuis2012spatially, nieuwenhuis2014co, santner2011interactive, vezhnevets2005growcut}, laid the foundation for this topic. Later, with the resurgence of deep learning, a few works (\eg, \cite{Agustsson_2019_CVPR,andriluka2020efficient}) have also contributed to the development of deep-learning-based methods for this problem.

%% file: sec/3_IOBE.tex
\vspace{-9pt}
\section{Methods}
\label{sec:methods}
\vspace{-6pt}




\subsection{Interactive Occlusion Boundary Estimation (IOBE)}
\label{sec:interaction_method}
\vspace{-6pt}

\textbf{~\interactionMechanism.}
Different from IS, IOBE aims to identify a set of boundaries that may be either open or closed.
Balancing effectiveness, efficiency, and simplicity, we propose the~\textbf{\interactionMechanism} mechanism, which constitutes the non-deep-learning component of the~\textbf{\interactiveMethodName~}framework.

In \interactionMechanism, \emph{boundary scribbles} enable users to mark misestimated boundary segments, where a \emph{FN-scribble} (False Negative scribble) or \emph{FP-scribble} (False Positive scribble) refers to a 2D connected region containing one or more FNs or FPs, respectively. In practice, these scribbles appear as curves that are significantly broader than actual OBs, and their wider form helps reduce the manual effort and attention required when inputting interactions.
Once the initial prediction is generated, the user provides multiple scribbles in a single step. The system then processes them in one shot and outputs the final refined result.

To integrate scribbles with the deep model, the FN and FP-scribbles are encoded into \emph{FN-map} and \emph{FP-map}, respectively. These maps are binary images with the same dimensions as the input image, where pixels have a value of 1 if they lie within at least one corresponding scribble and 0 otherwise. The concatenation of these maps forms the \emph{FN-FP map}, while their combined representation with the previous output constitutes the \emph{interaction map}.

\begin{figure*}[h]
 \vspace{-10pt}
  \centering
  \small
  \begin{minipage}{0.96\textwidth}
    \centering
    \includegraphics[height=0.2\textheight, width=\textwidth]{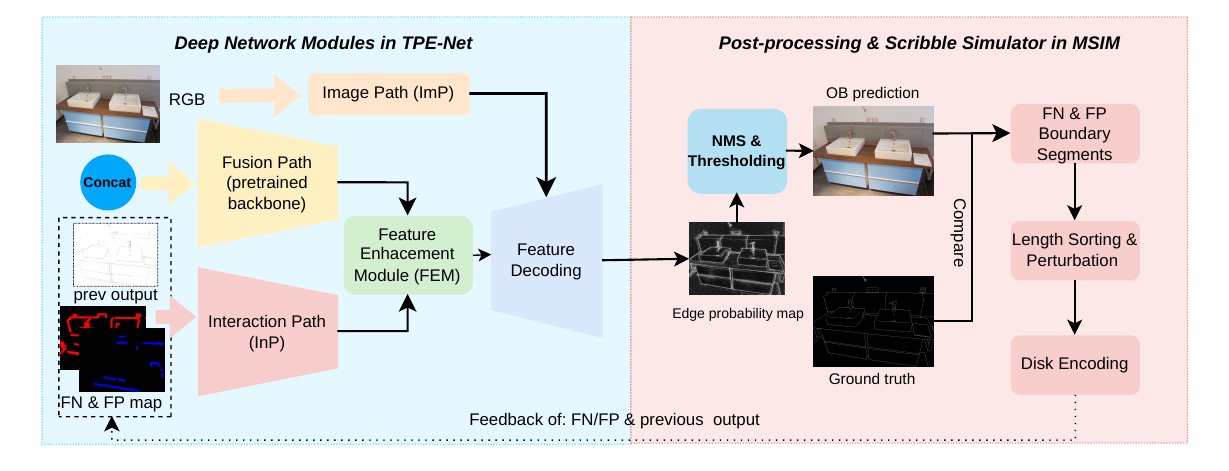} 
  \end{minipage}
  \vspace{1pt}
  \caption{\textbf{\interactiveMethodName~framework}: {\footnotesize integration of the deep network \deepModelName~into the interaction mechanism \interactionMechanism, which incorporates the boundary scribble approach, one-shot interaction strategy, two-stage training scheme, scribble simulator, \etc. \interactiveMethodName~is modular in terms of both the deep network and the interaction mechanism.}}
  \vspace{-8pt}
  \label{fig.model_pipeline}
\end{figure*}

\par\noindent\textbf{{\deepModelName.}} 
The proposed network employs a fusion path, based on a pretrained backbone, to encode the concatenated 6-channel input comprising the RGB image and the interaction map, following a common design in IS. To provide richer low-level details for the decoding process, and to make the model more aware of fine structures, we adopt an additional RGB image path—an approach shown to be effective in edge detection and OB estimation (\eg,~\cite{feng2021mt,pu2021rindnet})—thereby improving final boundary prediction.  
Since interaction features are much sparser than image features yet contain high-level cues such as spatial position, we introduce an \emph{interaction path} module that separately encodes the interaction map into deep hierarchical interaction features. This design corresponds to the \emph{late fusion mechanism} used in IS methods (\eg, \cite{Hao_2021_ICCV_edgeflow, hu2019fully}).
Together, these components form a three-path encoding design for \deepModelName.
In addition, to better incorporate the long, narrow scribble interaction features from the interaction path—and to enhance the model’s ability to capture long, narrow OBs—we adopt a multi-scale strip convolutional module, named Feature Enhancement Module (FEM), with receptive fields ranging from small to large in orthogonal directions, applied before the decoding process (see \emph{Supp.} for more details). 
\deepModelName's detailed workflow into \interactionMechanism~are illustrated in \reffig{fig.model_pipeline} and \emph{Overall Pipeline} below.

\vspace{3pt}
\par\noindent\textbf{{Overall Pipeline}} is illustrated in~\reffig{fig.model_pipeline}\footnote{In the paper, red \& blue colorizations are employed to provide a clear illustration of the FN \& FP maps.}: {
\textbf{(i)}~Three separate paths are proposed for encoding features: \emph{Fusion Path} encodes the six-channel input (consisting of RGB image, FN-FP map, and previous output) into hierarchical backbone features; \emph{Interaction Path} (InP) and \emph{Image Path} (ImP) separately process the interaction map and the RGB image, generating hierarchical interaction features and supplementary image information for use in FEM and \emph{Feature Decoding}, respectively;
\textbf{(ii)}~\emph{Fusion Path} backbone features are element-wise summed with interaction features. The resulting features are further deeply merged and enhanced by our designed FEM, which primarily consists of multi-scale strip convolution layers;
\textbf{(iii)}~\emph{Feature Decoding} combines the enhanced features with the low-level \emph{Image Path} features to produce the edge probability map;
\textbf{(iv)}~A non-maximum suppression (NMS) process is applied to the edge probability map to obtain a thin edge map, commonly used in OB estimation~\cite{wang2016doc, qiu2020pixel, feng2021mt}.} Then, \emph{thresholding} is done on the thin edge map to produce the OB prediction.
\textbf{(v)} In the training process and IOBE performance evaluation, for machine-simulated interactions, \interactiveMethodName~compares the GT with the model's previous output to automatically generate FN and FP scribbles. These scribbles are then perturbed, selected, and encoded as disks to form the FN-FP map, which is subsequently fed into the deep network~\deepModelName. 

\vspace{3pt}
\par\noindent\textbf{Training.} 
Unlike conventional IS methods, our~\interactionMechanism~introduces a novel two-stage training strategy. \textbf{The first stage} generates training samples by randomly sampling FN segments from GT boundaries and extracting FP segments from non-overlapping Canny edges. These segments are transformed into FN/FP-scribbles via our \emph{scribble simulation} method (detailed below). The network then trains on RGB images paired with dynamically initialized FN-FP maps (containing sparse scribbles or zeros) and zero-initialized previous outputs. Through a single forward pass, this stage simultaneously achieves: (1) autonomous OB prediction without interactive guidance, and (2) preliminary learning of scribble-to-error-region correlations - establishing \interactionMechanism’s adaptive foundation. \textbf{The second stage} implements an iterative refinement process through two sequential forward passes: (i) initial predictions are generated using zero-initialized FN-FP maps and previous outputs, which are then compared with ground truth to identify error boundary segments. The system selects \tc{the longest and most significant} FN/FP segments for \emph{scribble simulation} to produce updated FN-FP maps. (ii) the model performs retraining using these updated maps along with the original RGB images and the initial predictions (serving as the previous output), enabling one-shot interaction refinement through self-error correction, which significantly enhances prediction precision.

\vspace{3pt}
\par\noindent\textbf{Testing.}
For IOBE \textbf{performance evaluation}, we implement a machine-simulated interaction framework following standard IS protocols. The prediction process involves: (1) an initial phase generating both the initial prediction and machine-simulated FN-FP maps (matching training step (i)), followed by (2) a refinement phase combining the input image with its FN-FP map and the output obtained in phase (1). For \interactiveMethodName-based \textbf{annotation evaluation}, we replace the simulated scribble generation in phase (1) with actual users' FN/FP-scribbles while maintaining the same prediction process in phase (2).

\vspace{3pt}
\par\noindent\textbf{Scribble Simulation.} 
We simulate the FN/FP-scribbles and get the FN-FP map, by (i) introducing perturbations in both length and location to the set of FN/FP boundary segments and then (ii) using a disk kernel to dilate the resulting point sets.
The length perturbation is done by randomly lengthening/shortening the boundaries. And the position perturbation is applied after length perturbation, wherein each pixel point in the boundary segments receives an individual random perturbation of maximal extent.
Regarding (ii), it can be simply implemented by applying the disk encoding (\ie,~\cite{sofiiuk2022reviving}) to the resulting point sets. 

\vspace{-7pt}
\subsection{Synthetic Benchmark Generation}
\label{sec:occlusion_genertion}
\vspace{-3pt}
Given mesh-based 3D scene data and a camera model, we first generate an RGB image using Blender~\cite{blender_tool}, then employ~\sytheticGenerationName~to produce the corresponding OB map.

\begin{figure}[h]
  \vspace{-10pt}
  \centering
  \small
  \begin{minipage}{0.15\textwidth}
    \centering
    \includegraphics[width=\textwidth]{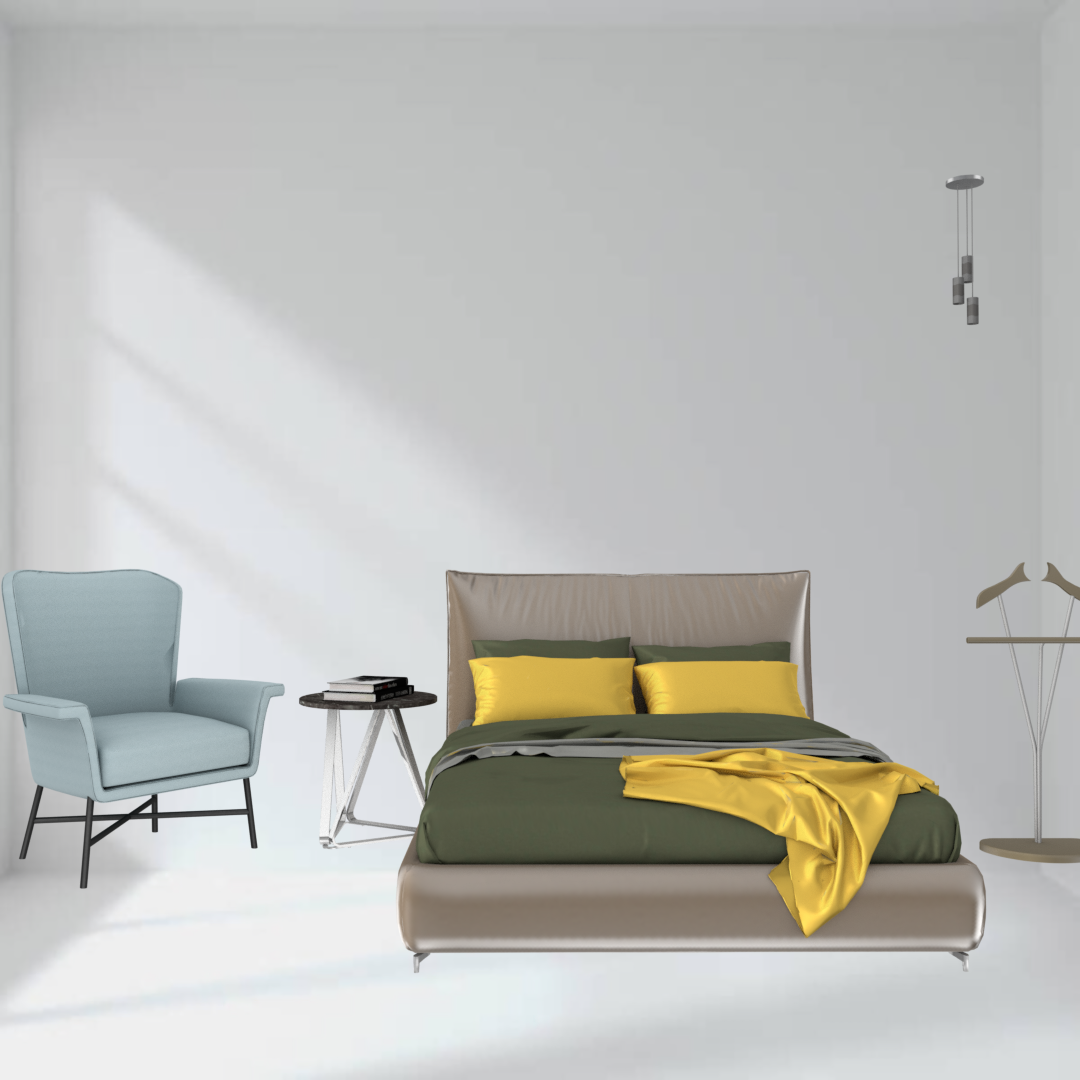}
  \end{minipage}
  \hspace{-3pt}
  \begin{minipage}{0.15\textwidth}
    \centering
    \includegraphics[width=\textwidth]{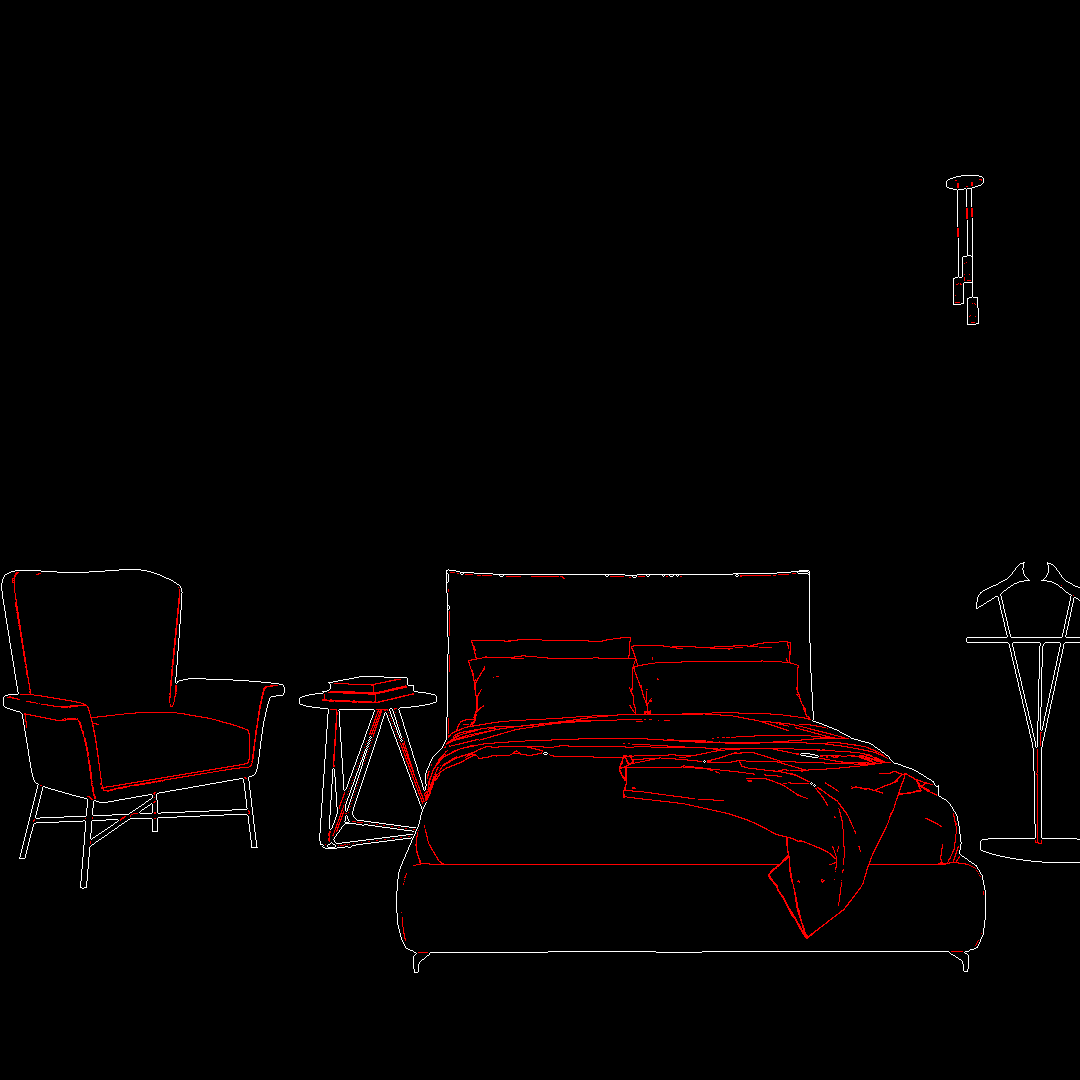}
  \end{minipage}
  \hspace{1pt}
  \begin{minipage}{0.15\textwidth}
    \centering
    \includegraphics[width=\textwidth]{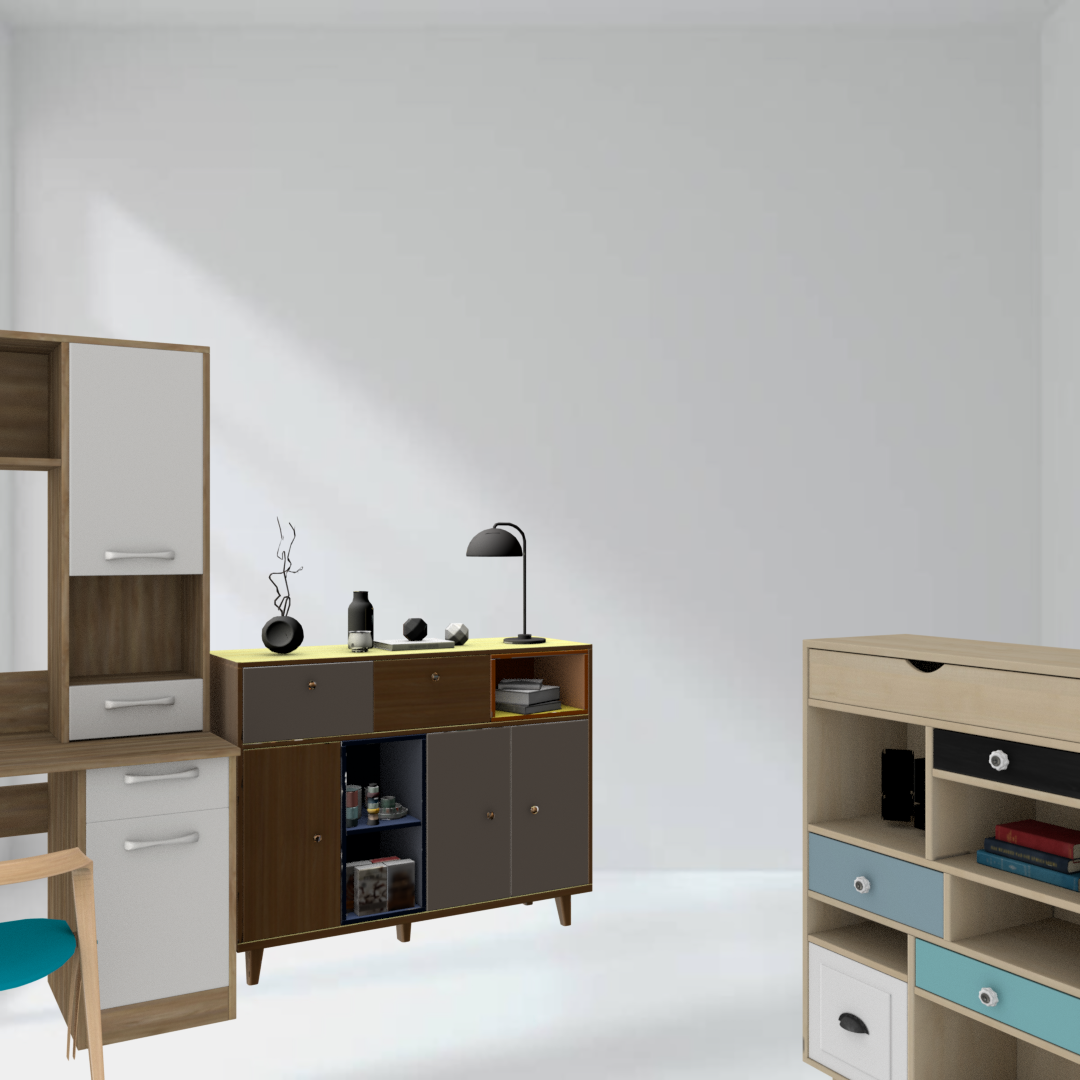}
  \end{minipage}
  \hspace{-3pt}
  \begin{minipage}{0.15\textwidth}
    \centering
    \includegraphics[width=\textwidth]{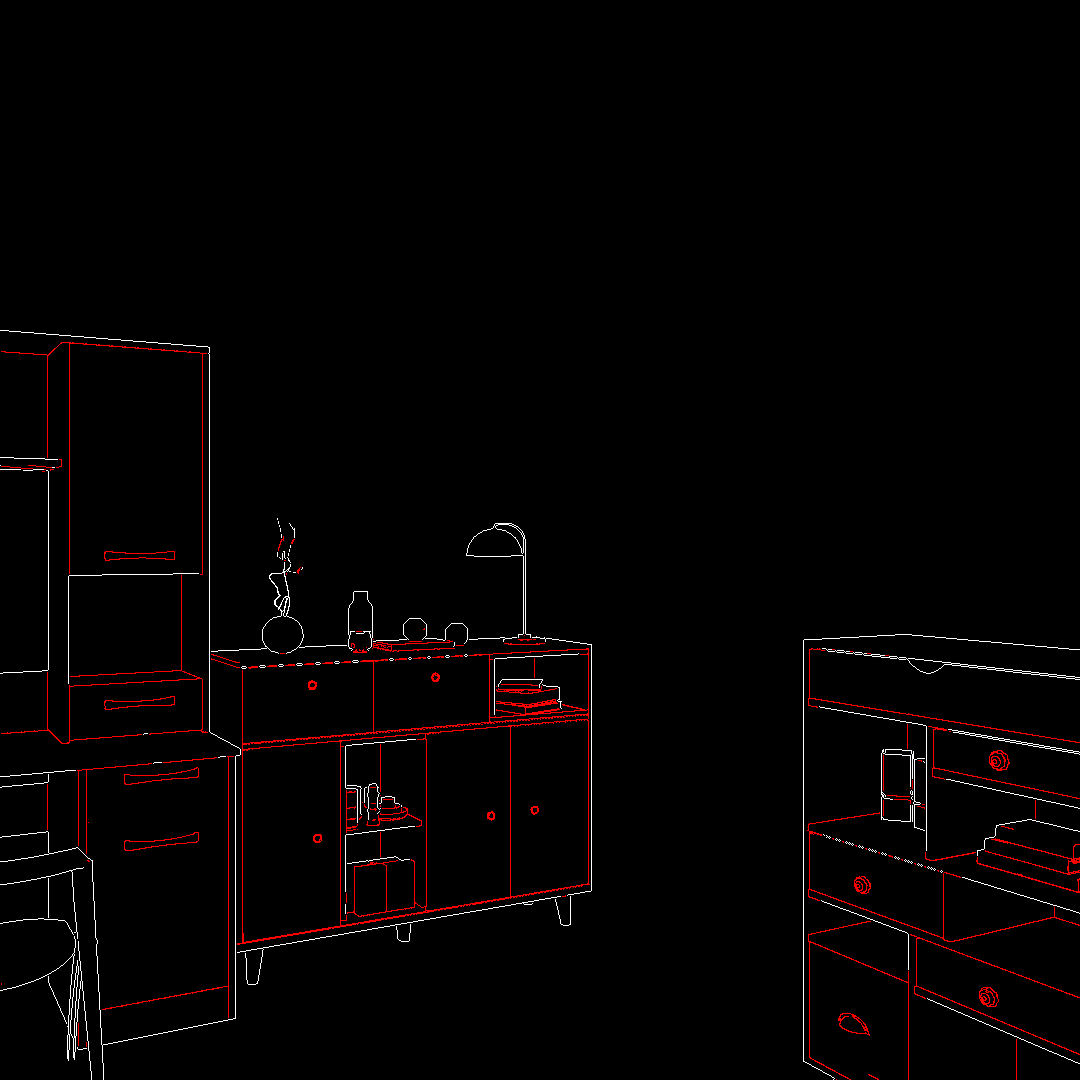}
  \end{minipage}
  \hspace{1pt}
  \begin{minipage}{0.15\textwidth}
    \centering
    \includegraphics[width=\textwidth]{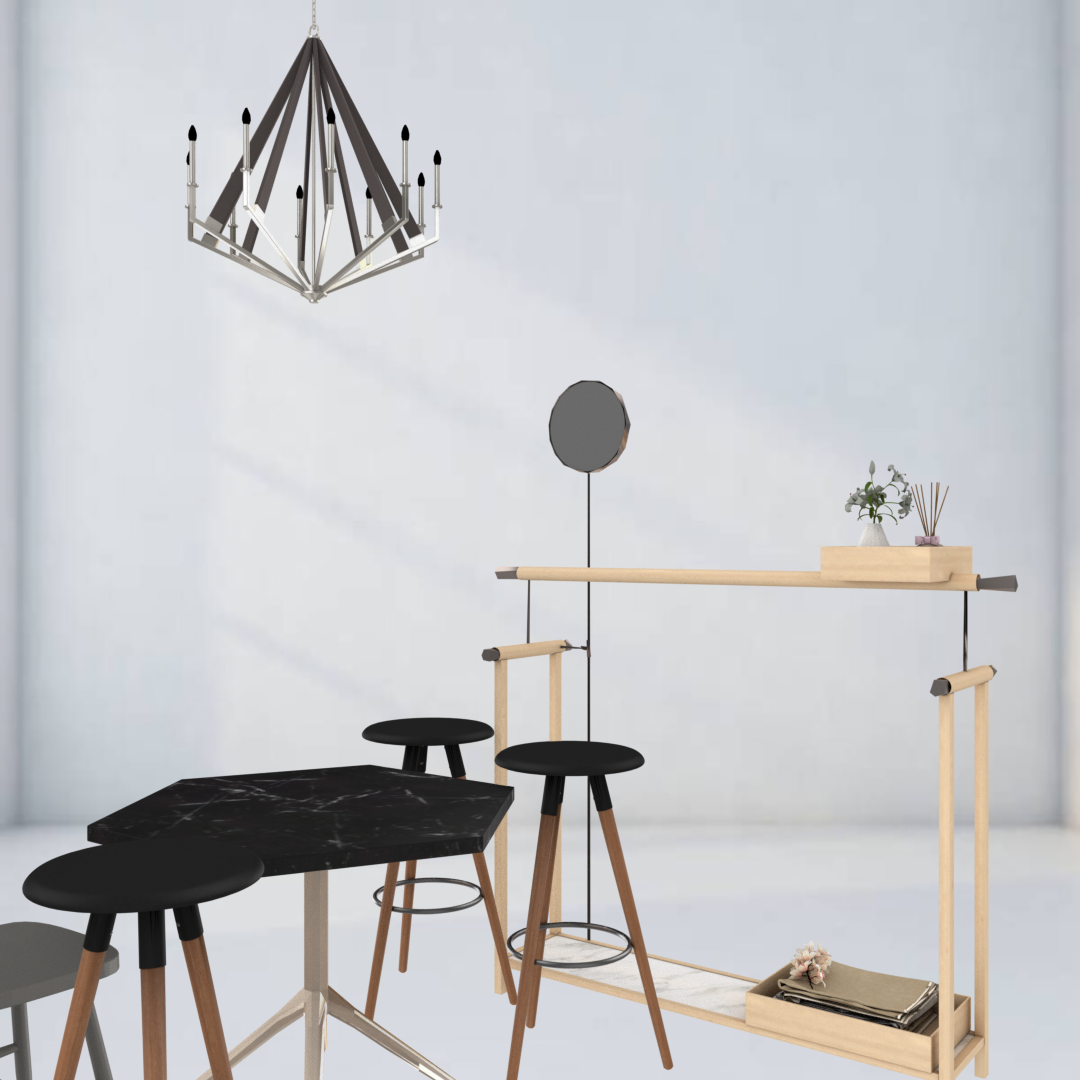}
  \end{minipage}
  \hspace{-3pt}
  \begin{minipage}{0.15\textwidth}
    \centering
    \includegraphics[width=\textwidth]{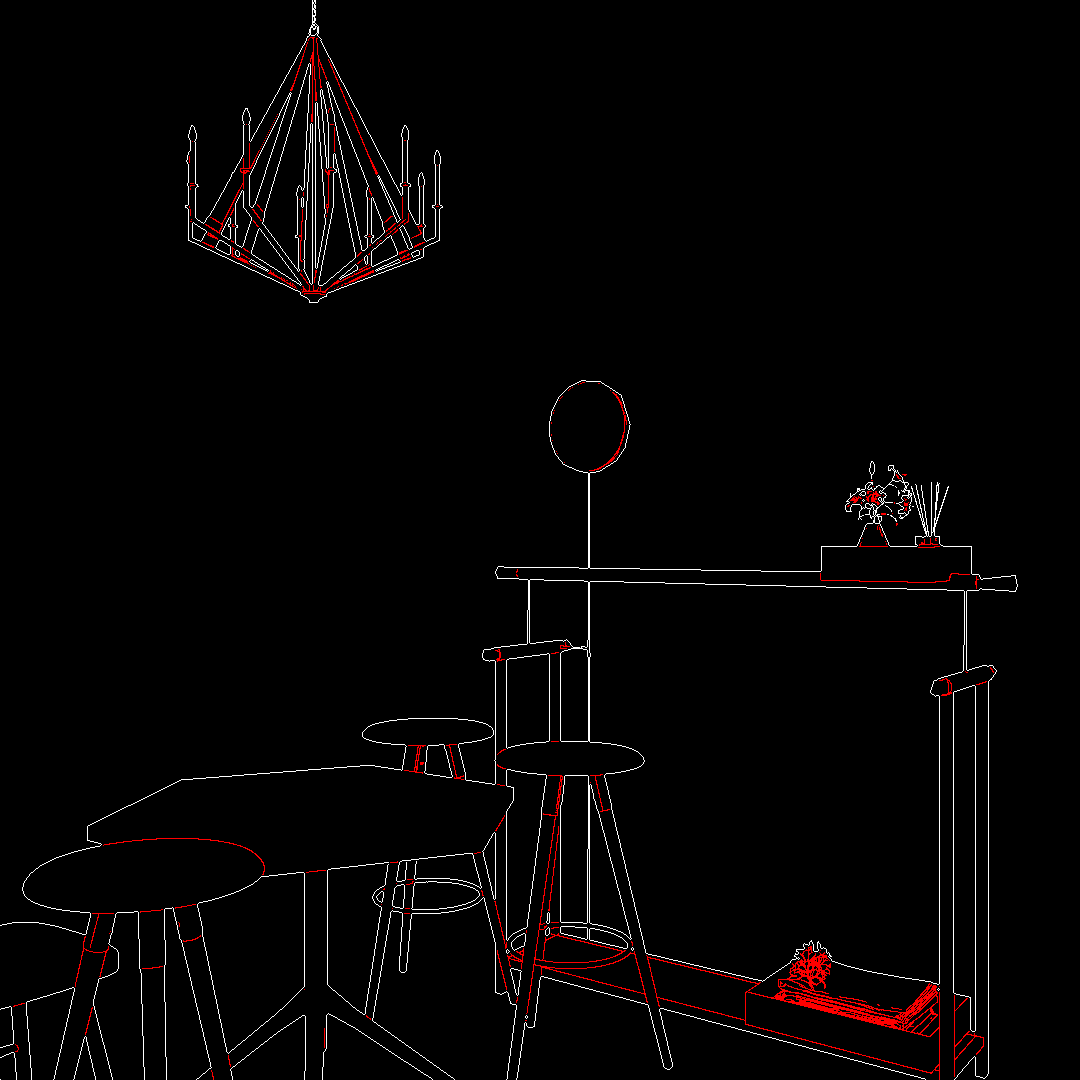}
  \end{minipage}
  \vspace{6pt}
  \caption{\textbf{Comparison of OB generation quality}: ~\sytheticGenerationName~vs.~\cite{qiu2020pixel}. \sytheticGenerationName's result: the union of those OBs in white \& \good{red};~\cite{qiu2020pixel}'s result: only those OBs in white.}
  \vspace{-10pt}
  \label{fig.synocc_compare_p2orm}
\end{figure}

\sytheticGenerationName~is a Python-based tool that implements the OB definition from~\cite{wang2020occlusion}, accurately generating both object-level and self-occlusion boundaries.
It identifies OB points via per-pixel classification, labeling a pixel as an OB point when an occlusion event occurs between its corresponding mesh face and the mesh face of one neighboring pixel. Specifically, we build a visibility map for 3D mesh faces using a graph-based data structure. For each pair of pixels being checked, we determine occlusion by evaluating the visibility path between their corresponding mesh faces, considering all intermediate faces along the path.
The Complete implementation details, more discussions and limitations are provided in \emph{Supp.} due to space limitations.
\reffig{fig.synocc_compare_p2orm} presents a qualitative comparison of OB generation between~\sytheticGenerationName~and the method of~\cite{qiu2020pixel}, the only previous approach capable of handling self-occlusions from depth and normal maps.
This not only validates the rationale of deriving OBs for 2D images directly from 3D scene data, but also demonstrates~\sytheticGenerationName's ability to capture both inter-object boundaries and fine-grained self-occlusion structures.

Note that the size of the generated~\sytheticBenchmarkName~dataset can be flexibly scaled by adjusting camera settings. Moreover, \sytheticGenerationName~supports the creation of unlimited high-quality OB datasets from diverse 3D sources beyond indoor \sytheticBenchmarkName, of which superior quality is demonstrated by high performance in all experiments without domain adaptation.

%% file: sec/4_experiments.tex
\vspace{-9pt}
\section{Experiments}
\label{sec:experiment}
\vspace{-7pt}

In this section, we sequentially present the experimental setup, the qualitative and quantitative results, and the ablation studies. More dataset details, experiments (\eg,~\interactiveMethodName~on more datasets, OB applications,~\etc), and additional ablation studies are provided in \emph{Supp}. 

\vspace{-6pt}
\subsection{Experimental Setup} 
\label{sec:experimentalSetup}

\vspace{-2pt}

\textbf{Datasets.}\footnote{
See~\refsec{sec:intro}~(Contributions) and Table~\ref{tab-db_compare}~for an overview of~\sytheticBenchmarkName~and~\realBenchmarkName. Additional details (\eg, \sytheticBenchmarkName’s construction and~\realBenchmarkName’s annotation procedures, \etc) are provided in \emph{Supp}.}
Unless otherwise specified, proposed \interactiveMethodName~was trained using the synthetic~\sytheticBenchmarkName, and the testing was conducted based on both~\subEntitySeg~and~\subDIODE.
For ablation studies, due to space limitations, we present only the~\subEntitySeg~results using the Swinformer base, and the~\subDIODE~results shown in \emph{Supp.} yield consistent conclusions.

\vspace{3pt}
\noindent\textbf{Implementation Details.} The main pretrained backbone is \emph{Swinformer architecture}~\cite{liu2021swin}, unless explicitly stated otherwise.
The disk's radius for performing the scribble simulation is $12$.
The first training stage is conducted over $4$ epochs, and the second stage is started from epoch $5$.
More detailed training hyperparameters are stated in \emph{Supp}. 
During training, the maximum allowable number of the longest candidate FN/FP boundary segments is set to 5.
For IOBE performance evaluation (\ie, machine-simulated one), all the FN/FP boundary segments whose length is larger than 30 are taken to generate FN/FP-scribbles.
Following OPNet~\cite{feng2021mt}, we use the class-balanced cross-entropy loss to supervise the training process.

\vspace{3pt}
\par\noindent\textbf{Baselines.}  
Due to the lack of prior work on IOBE, we adapted baselines from seven state-of-the-art IS methods~
\cite{lin2020fcanet,liew2021deepthin,sofiiuk2022reviving,chen2021conditional,chen2022focalclick,liu2023simpleclick,lin2025adapclick}
for IOBE performance comparison. The comparison with these IS competitors focuses on two respects: the network architecture and the interaction mechanism. The conventional IS Interaction Mechanism (\textbf{ISIM}) differs from our~\textbf{\interactionMechanism}~in three key aspects: the use of boundary scribbles, one-shot interaction strategy, and a two-stage training scheme. As discussed in Section~\ref{sec:related_work}, our intuitive boundary scribbles are conceptually distinct from the point- or region-based clicks/scribbles commonly used in IS methods, which typically target single-object segmentation. To compare against the region-based click interaction approach, we adopt the \emph{boundary clicks} paradigm, where users correct FN/FP errors by clicking near the center of erroneous boundary segments. Unlike~\interactionMechanism, conventional ISIM follows a one-stage training scheme with progressive multi-round refinement, where results may not consistently improve across iterations. Specifically, in each round, either an FN- or FP-interaction is generated based on the most significant error; then the previous output will be updated, and the FN-FP map will be refined by merging prior and new interactions, which are then used to predict the next-round result. Each iteration has a runtime approximately equal to that of~\interactiveMethodName. Unless otherwise specified, we report ISIM results based on 10 rounds of refinement.
Extensive experiments demonstrate the superiority of~\deepModelName~and the three key components of \interactionMechanism~for efficiently handling full-image OBs in IOBE. Additionally, we evaluated OPNet~\cite{feng2021mt}, a SOTA fully automatic method, to provide a baseline for measuring the improvements gained through interactions. 

\noindent\textbf{Evaluation Modes.} As discussed in \refsec{sec:methods}, \interactiveMethodName~uses two testing modes. Most experiments are evaluated under the IOBE performance evaluation setting (\ie, machine-simulated interactions) following IS protocols. However, to demonstrate the framework’s applicability, Table~\ref{tab-time_cost} presents a user study in which human annotators manually draw FN–FP scribbles offline by inspecting the previous OB predictions and applying their knowledge to judge true OBs. These interaction maps are then fed to the model to obtain refined outputs.

%
\noindent\textbf{Evaluation Metrics.} \emph{Precision} \& \emph{Recall} (PR) are calculated using the same protocol as those closely related works~\cite{feng2021mt, qiu2020pixel, lu2019occlusion, wang2019doobnet, wang2016doc}. Furthermore, the following 3 standard metrics are calculated from PR: (i) \emph{Fixed contour threshold (ODS)}, which is the F-measure with the best fixed OB probability threshold over the all datasets; (ii) \emph{Best threshold of image (OIS)}, which is F-measure with the best OB probability threshold for each image; (iii) \emph{Average precision (AP)}, which is the average precision over all occlusion probability thresholds.

\begin{figure}[htbp]
    \vspace{-8pt}
    \centering
    \small
    \resizebox{0.99\textwidth}{!}{
        \begin{tabular}{c}
            \hspace{1em} RGB Image
            \qquad Prev Output
            \qquad\quad FN-FP Map
            \qquad Final Output
            \qquad Ground Truth
            \vspace{1pt}\\

            \includegraphics[width=0.18\linewidth]{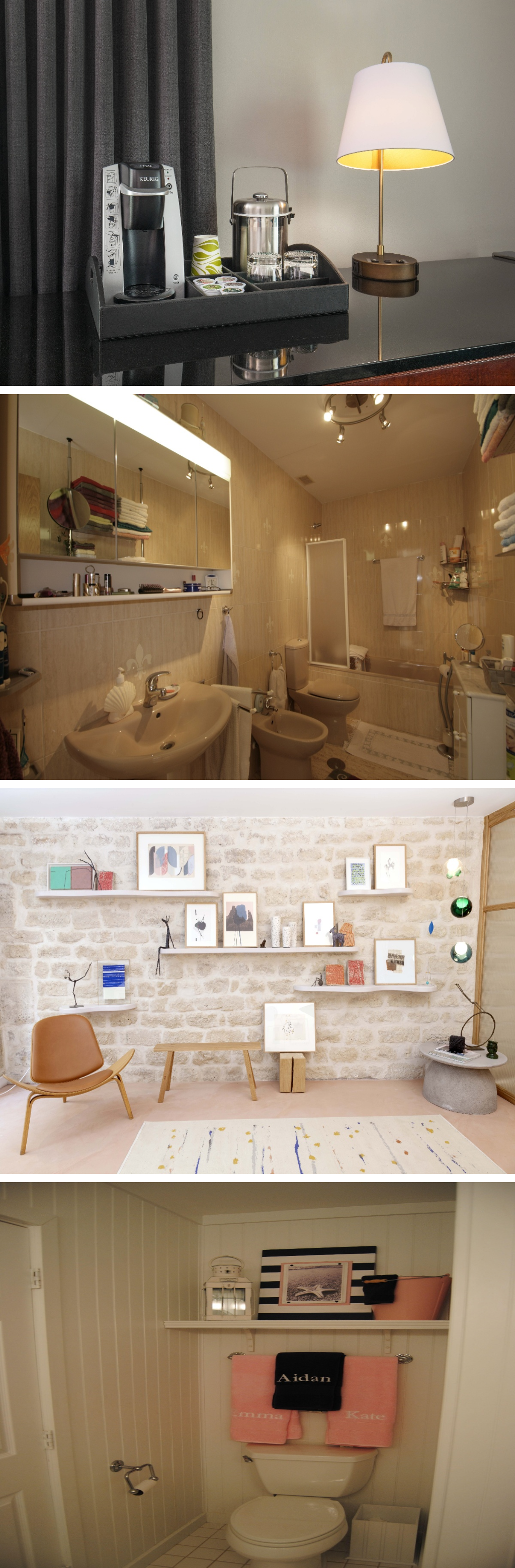}
            \includegraphics[width=0.18\linewidth]{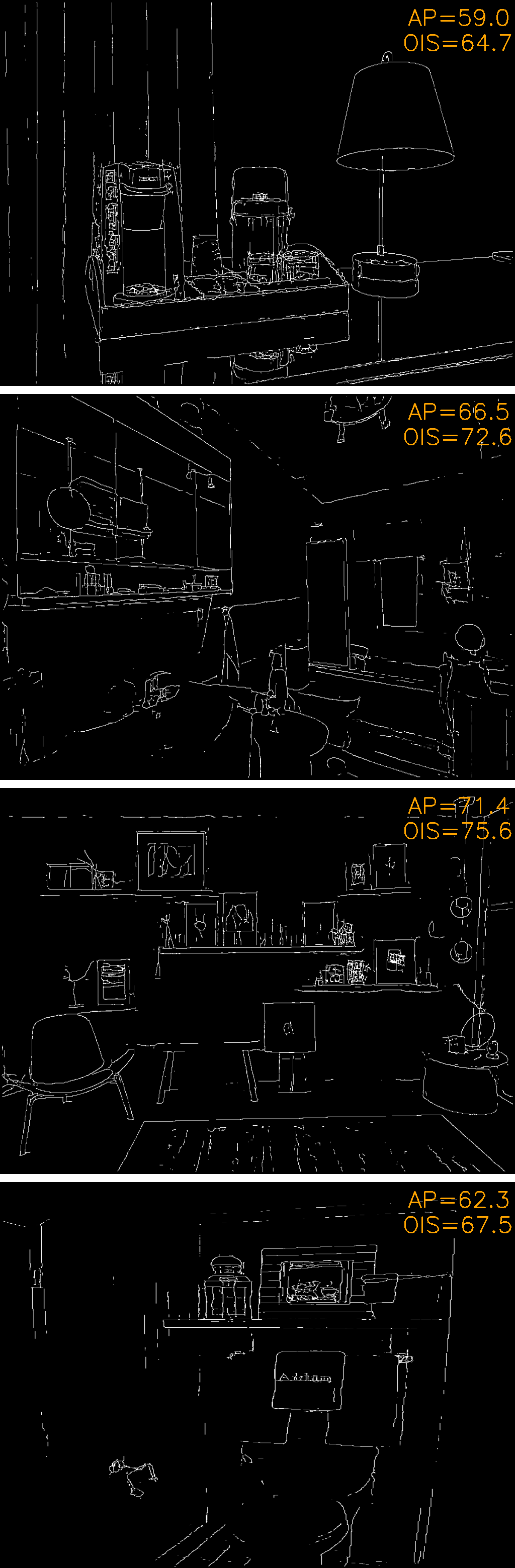}
            \includegraphics[width=0.18\linewidth]{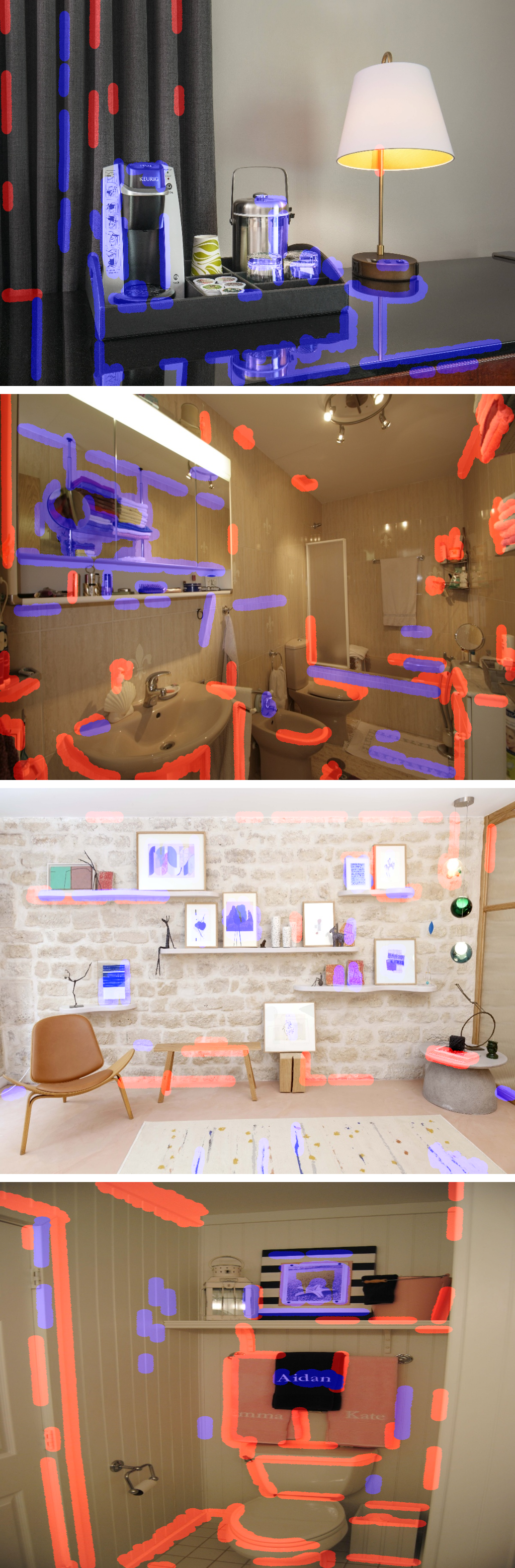}
            \includegraphics[width=0.18\linewidth]{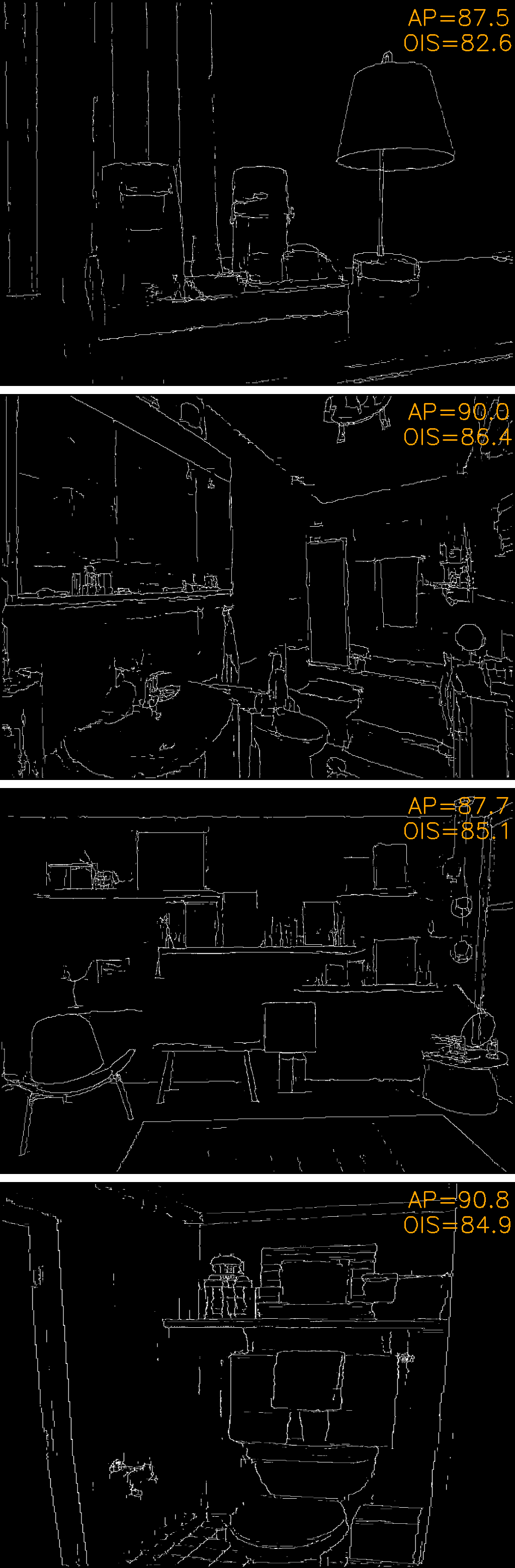}
            \includegraphics[width=0.18\linewidth]{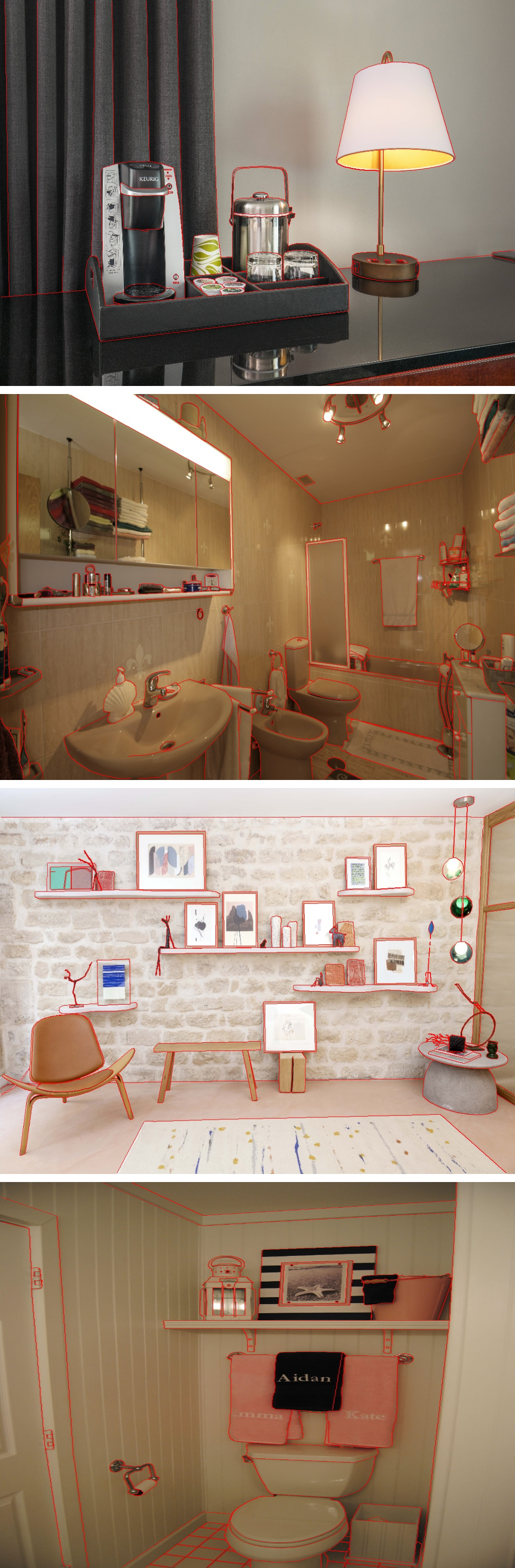}
            \\
        \vspace{-5pt}
        \end{tabular}
    }
    \caption{\small\textbf{Qualitative results} with quantitative metrics displayed in top-right corners (columns 2\&4)}
    \label{fig.vis_iobe}
    \vspace{-8pt}
\end{figure}

\vspace{-6pt}
\subsection{Qualitative \& Quantitative Results}
\vspace{-3pt}

\reffig{fig.vis_iobe}~presents qualitative results of~\interactiveMethodName~on four representative examples. The framework demonstrates strong performance, with boundary scribble interactions substantially enhancing prediction quality. 
Table~\ref{tab-time_cost} reports annotation times and quality assessment metrics from experiments on 20-sample subsets of~\subDIODE~and \subEntitySeg. The quantitative results demonstrate~\interactiveMethodName's effectiveness for OB annotation with human-provided scribbles, highlighting the utility of interactive estimation in annotation workflows.

\begin{table}[htbp]
    \begin{minipage}[t]{0.48\textwidth}
    \vspace{-5pt}
    \tiny
    \centering
    \caption{\textbf{\footnotesize Human annotation time \& quality of~\interactiveMethodName}.
    \footnotesize Fully manual annotation takes $6 \sim 10$ times longer and requires significantly more meticulous effort compared to \interactiveMethodName-based annotation.}
    \label{tab-time_cost}
    \vspace{5pt}
    \begin{tabular}{lccccc}
        \toprule
        \textbf{Dataset} & \textbf{Time cost$\downarrow$} (mins)  & \textbf{ODS$\uparrow$} & \textbf{OIS$\uparrow$} & \textbf{AP$\uparrow$} \\
        \midrule
        \textbf{\subDIODE~}& \textbf{5\textpm1} & \textbf{88.8} & \textbf{89.2} & \textbf{90.7} \\
        \midrule
        \textbf{\subEntitySeg~}& \textbf{6\textpm1.5} & \textbf{84.8} & \textbf{86.1} & \textbf{86.8} \\
        \bottomrule
    \end{tabular}

    \vfill
    \vspace{3pt}
    
    \tiny
    \centering
        \caption{\footnotesize {\textbf{Analysis on the component benefits of~\interactiveMethodName}}.
        \dag{} indicates that the corresponding model was trained using the conventional IS Interaction Mechanism (ISIM).
        Optimal results per block or overall are highlighted with \textbf{bold} or \textbf{\underline{underlines}}, respectively. These conventions are applied throughout the remainder of this work unless otherwise specified.
       }
        \label{tab-ablation_method_component}
        \tiny
        \vspace{4pt}
        \begin{tabular} {ccccccc} 
            \toprule 
            \textbf{InP} & \textbf{ImP} & \textbf{FEM} & \textbf{Interaction} & \textbf{ODS$\uparrow$} & \textbf{OIS$\uparrow$} & \textbf{AP$\uparrow$} \\
            \midrule
            $\mathbf{\times}$ & $\mathbf{\times}$ & $\mathbf{\times}$ & scribbles  & 48.8 & 50.2 & 41.8 \\ 
            $\mathbf{\times}$ & $\checkmark$ & $\mathbf{\times}$  & scribbles & 76.1 & 78.1 & 80.6 \\
            $\checkmark$ & $\checkmark$ & $\mathbf{\times}$ & scribbles & 72.9 & 75.4 & 77.4 \\ 
            $\mathbf{\times}$ & $\checkmark$ & $\checkmark$  & scribbles & \textbf{80.0} & \textbf{81.2} & \textbf{85.4} \\
            \midrule
            $\checkmark$ & $\checkmark$ & $\checkmark$ & clicks \dag{} & 46.1 & 52.8 & 39.5 \\ 
            $\checkmark$ & $\checkmark$ & $\checkmark$ & scribbles \dag{} & 43.5 & 52.0 & 36.7 \\
            $\checkmark$ & $\checkmark$ & $\checkmark$ & clicks & 71.4 & 73.4 & 73.9 \\ 
            $\checkmark$ & $\checkmark$ & $\checkmark$  & scribbles & \textbf{\underline{81.3}} & \textbf{\underline{82.5}} & \textbf{\underline{86.4}}  \\  
            \bottomrule 
        \end{tabular}
        
    \end{minipage}
    \hfill
    \begin{minipage}[t]{0.5\textwidth}
        \vspace{-5pt}
        \centering
        \tiny
        \caption{\footnotesize \textbf{Quantitative comparison} between the fully automatic OPNet, IS-based baselines, and our~\interactiveMethodName. We present the results of IS methods trained and tested using the conventional ISIM with \emph{boundary click} in the second block, marked with \ddag{}, while the third block presents the performance of the same IS methods trained with our proposed interaction mechanism~\interactionMechanism.}

      \label{tab-main_s12}

      \vspace{5pt}
      \begin{tabular} {lccc|ccc} 
      \toprule 
      \multicolumn{1}{c}{} & \multicolumn{3}{c}{\textbf{\subDIODE}} & \multicolumn{3}{c}{\textbf{\subEntitySeg}} \\
      \cmidrule(lr){2-4} \cmidrule(lr){5-7}

      \textbf{Method}  & \textbf{ODS$\uparrow$} & \textbf{OIS$\uparrow$} & \textbf{AP$\uparrow$}  & \textbf{ODS$\uparrow$} & \textbf{OIS$\uparrow$} & \textbf{AP$\uparrow$} \\

      \midrule 
      OPNet~\cite{feng2021mt} & 72.8 & 74.0 & 61.8  & 50.7 & 57.7 & 36.8 \\

      
      \midrule 
      RITM~\cite{sofiiuk2022reviving} \ddag{}  & 61.9 & \textbf{70.3} & 55.6 & 24.6 & 30.1 & 11.3 \\
      CDNet~\cite{chen2021conditional} \ddag{} & 35.1 & 44.5 & 22.2 & 34.3 & 38.3 & 24.3 \\
      TOS-Net~\cite{liew2021deepthin} \ddag{} & 58.4 & 68.7 & 46.1 & \textbf{57.5} & \textbf{60.3} & \textbf{50.4} \\
      FCA-Net~\cite{lin2020fcanet} \ddag{} & 58.9 & 66.5 & 55.7 & 49.7 & 52.2 & 40.3 \\
      FocalClick~\cite{chen2022focalclick} \ddag{}  & 60.6 & 68.9 & \textbf{59.4} & 47.6 & 51.8 & 44.5 \\
      AdaptiveClick~\cite{lin2025adapclick} \ddag{} & 61.3 & 69.1 & 55.8 & 47.8 & 49.5 & 39.2 \\      
      SimpleClick~\cite{liu2023simpleclick} \ddag{} & \textbf{64.4} & 69.4 & 59.3 & 50.7 & 54.3 & 44.0 \\
      
      \midrule       

      RITM~\cite{sofiiuk2022reviving}  & 75.0 & 78.0 & 76.0  & 32.8 & 35.1 & 13.7  \\
      CDNet~\cite{chen2021conditional}   & 67.6 & 71.2 & 65.6  & 43.2 & 50.7 & 35.3  \\
      TOS-Net~\cite{liew2021deepthin}  & 74.8 & 81.8 & 60.4  & 69.0 & 70.9 & 54.9  \\
      FCA-Net~\cite{lin2020fcanet}  & \textbf{77.1} & \textbf{82.6} & \textbf{81.9}  & \textbf{71.8} & \textbf{75.2} & \textbf{73.7}  \\
      FocalClick~\cite{chen2022focalclick}   & 73.9 & 79.0 & 76.9  & 63.3 & 69.0 & 68.6 \\
      AdaptiveClick~\cite{lin2025adapclick}   & 75.8 & 79.8 & 79.0  & 66.2 & 69.7 & 64.2  \\
      SimpleClick~\cite{liu2023simpleclick}  & 73.6 & 78.5 & 78.2  & 69.5 & 71.4 & 70.0  \\
      \midrule 

      \textbf{Ours}  Swin-S & 84.2 & 87.0 & 89.4  & 81.2 & 82.1 & \textbf{\underline{86.4}}  \\
      \textbf{Ours}  Swin-B & 85.5 & \textbf{\underline{87.9}} & 90.0 & \textbf{\underline{81.3}} & \textbf{\underline{82.5}} & \textbf{\underline{86.4}} \\

      \textbf{Ours} Swin-L & \textbf{\underline{86.3}} & 87.8 & \textbf{\underline{91.4}}  & 79.3 & 80.9 & 84.8  \\
      
      \bottomrule 
        \end{tabular}
    \end{minipage}
    \vspace{-7pt}
\end{table}

Table~\ref{tab-ablation_method_component} shows ablation results for the proposed \interactionMechanism~and \deepModelName~components in~\interactiveMethodName.
In the first block, we evaluate four architectural variants by removing InP, ImP, and/or FEM. The quantitative results demonstrate that the complete model (last row, second block in Table~\ref{tab-ablation_method_component}) achieves optimal performance when all components are integrated.
In the second block, we evaluate four interaction paradigms using the full~\deepModelName: conventional ISIM with boundary clicks and scribbles, and our \interactionMechanism~with boundary clicks and scribbles. The ISIM variants mainly differ in the one-shot interaction and two-stage training scheme. The results show that \interactionMechanism~significantly outperforms compared IS mechanisms, with boundary scribbles yielding better performance than boundary clicks under \interactionMechanism~setting.
These results validate both~\deepModelName's network design and the proposed \interactionMechanism~for IOBE.

Table~\ref{tab-main_s12} presents a quantitative comparison between~\interactiveMethodName, the fully automatic OPNet~\cite{feng2021mt}~and IS-based competitors using ISIM or our \interactionMechanism. Our \interactiveMethodName~achieves the highest ODS, OIS, and AP scores on both \subDIODE~and \subEntitySeg, significantly outperforming the IS-based competitors by a large margin (\eg, on~\subEntitySeg~\interactiveMethodName~achieves improvements of over +9.5 ODS, +7.3 OIS, and +12.7 AP). 
In addition, \interactiveMethodName~shows substantial improvements over the fully automatic OPNet (\eg, ODS +30.6, OIS +24.8, and AP +49.6 on~\subEntitySeg, demonstrating the benefit of interaction in OB estimation.
The second and third blocks of the table present the performance of IS baselines using boundary clicks with the conventional ISIM and boundary scribbles with our proposed \interactionMechanism, respectively.
Two key observations emerge: (i) \interactionMechanism~with boundary scribbles is clearly more suitable for IOBE, as further evidenced by the performance improvements of IS methods between the second and third blocks. (ii) under the same \interactionMechanism~setting, our \deepModelName~(in fourth block) demonstrates higher utilization efficiency of scribble interactions, achieving better performance than IS methods. 
In \emph{Supp.}, we provide additional details in Table~\ref{tab-main_s12}, including metrics before interaction and other supporting data, to further validate our results.
These findings further highlight both the effectiveness of our \deepModelName~design and the advantages of the proposed \interactionMechanism~in \interactiveMethodName~framework.

\begin{wrapfigure}[11]{r}{0.5\textwidth}
    \vspace{-10pt}
    \centering
    \setlength{\tabcolsep}{2pt} 
    \renewcommand{\arraystretch}{1.2}

    \begin{tabular}{@{}c c@{}}
        {\scriptsize\rotatebox{90}{\textbf{Image}}} & \includegraphics[width=0.42\textwidth]{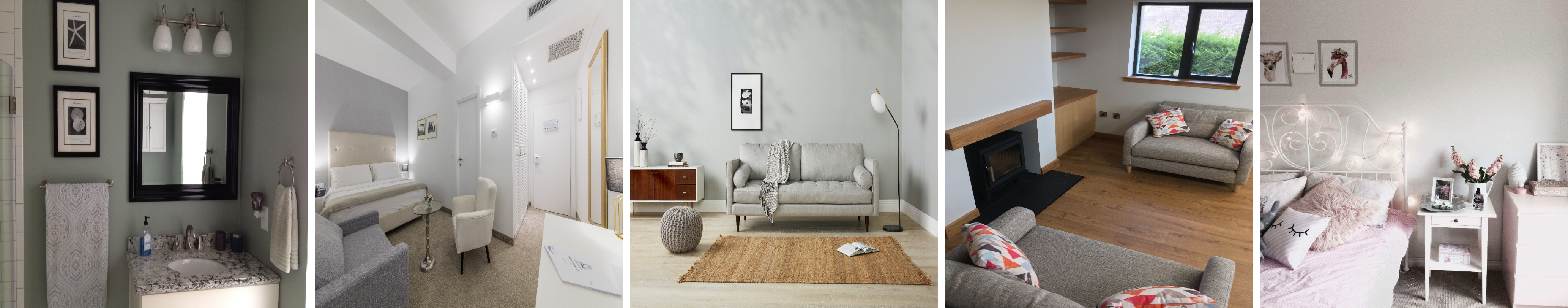} \\
        {\scriptsize\rotatebox{90}{\textbf{Before}}} & \includegraphics[width=0.42\textwidth]{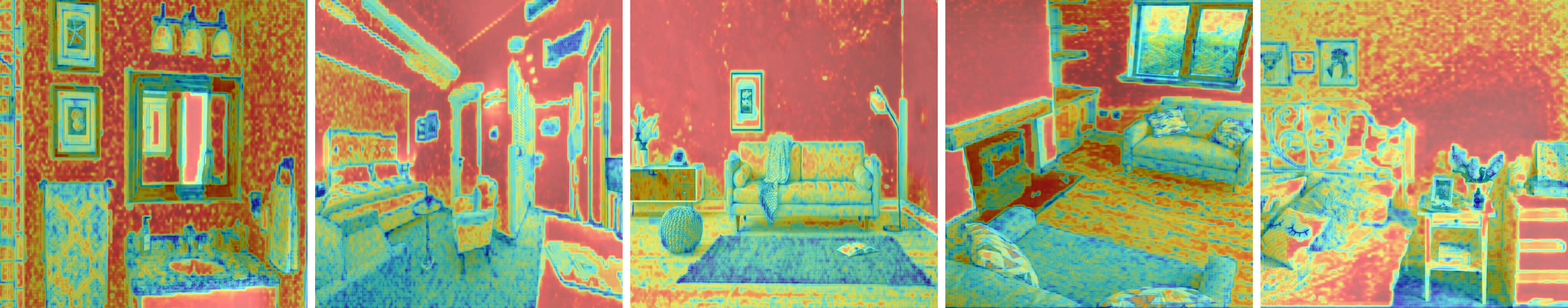} \\
        {\scriptsize\rotatebox{90}{\textbf{After}}} & \includegraphics[width=0.42\textwidth]{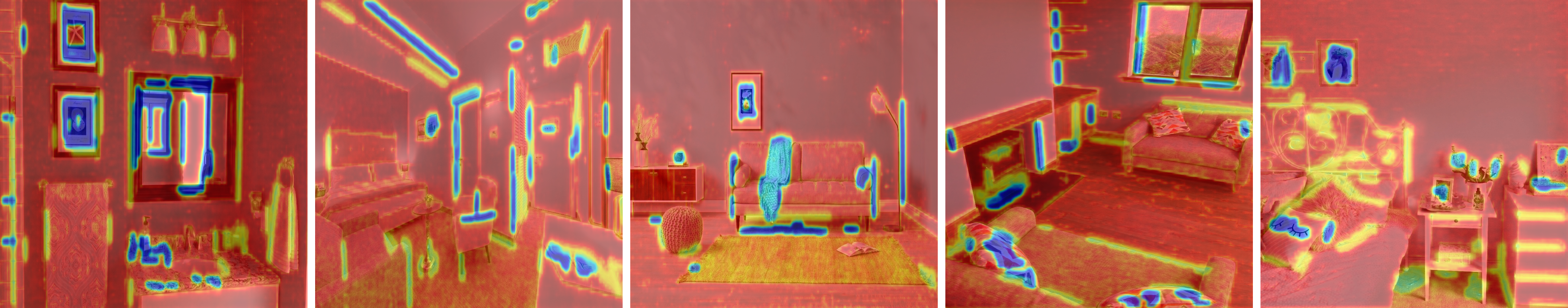}
    \end{tabular}

    
    \captionsetup{format=plain,justification=justified,width=0.48\textwidth}
    \vspace{5pt}
    \caption{\footnotesize {\textbf{Feature Heat Maps Before \& After FEM.} Given the feature maps from the Fusion Path and those enhanced by FEM, we perform channel-wise averaging to generate heat maps overlaid on the images.}}
    \label{fig.attnVis_FEM}
    \vspace{-13pt}
\end{wrapfigure}

Finally,~\reffig{fig.attnVis_FEM}~presents additional examples from~\realBenchmarkName, visualizing the feature enhancements produced by FEM. These results demonstrate that FN/FP boundary scribble signals become clearly highlighted via the proposed FEM, validating the rationale behind FEM's design.

\vspace{-5pt}
\subsection{Ablation Studies}
\vspace{-5pt}

Table~\ref{tab-ablation-prev-loop} presents results for two variants of~\interactiveMethodName: (i) removing the previous output from the model input, and (ii) replacing~\interactionMechanism~one-shot interaction approach with the progressive refinement interaction strategy from ISIM during both training and testing. As shown, the IS-style progressive interaction way not only increases computational cost but also yields in inferior performance, with potential degradation over multiple refinement rounds.

\begin{table}[htbp]
\tiny
    \begin{minipage}[t]{0.52\textwidth}
        \vspace{-7pt}
        \tiny
        \centering
        \caption{\textbf{\footnotesize {Ablation on~\interactiveMethodName: one-shot vs. progressive interaction and variant without previous output.}} 
        }  
        \label{tab-ablation-prev-loop}
        \vspace{3pt}
    \begin{tabular} {cccccc} 
        \toprule 
        \textbf{Progressive} & \textbf{Iteration} & \textbf{Prev Output} & \textbf{ODS$\uparrow$} & \textbf{OIS$\uparrow$} & \textbf{AP$\uparrow$}  \\
        
        $\checkmark$ & 1 & $\checkmark$  & 47.3 & \textbf{54.2} & 41.6 \\
        $\checkmark$ & 3 & $\checkmark$  & \textbf{47.9} & 53.3 &\textbf{42.3} \\
        $\checkmark$ & 5 & $\checkmark$  & 47.8 & 52.8 & 42.2 \\
        $\checkmark$ & 10 & $\checkmark$  & 47.7 & 52.6 & 42.1 \\
        \midrule
        $\mathbf{\times}$ & 1 & $\checkmark$  & \textbf{\underline{81.3}} & \textbf{\underline{82.5}} & \textbf{\underline{86.4}}  \\
        $\checkmark$ & 1 & $\mathbf{\times}$  & 75.3 & 76.8 & 79.2 \\
            \bottomrule 
        \end{tabular}
    \end{minipage}
    \hfill
    \begin{minipage}[t]{0.45\textwidth}
    \vspace{-7pt}
    \tiny
    \centering
    \caption{\textbf{\footnotesize Performance with different SSTS.}}
    \label{tab-abltation_SSIT}
    \vspace{7pt}
    \begin{tabular} {lccc} 
        \toprule 
        \textbf{SSTS} & \textbf{ODS$\uparrow$} & \textbf{OIS$\uparrow$} & \textbf{AP$\uparrow$} \\
        \midrule 
        1  & 38.8 & 45.3 & 30.0  \\
        \midrule
        2  & 71.6 & 73.9 & 76.6 \\
        3  & 77.3 & 78.8 & 82.8  \\
        4  & 79.4 & 80.5 & 83.5  \\
        5  & \textbf{81.3} & \textbf{82.5} & \textbf{86.4}  \\
        6  & 78.1 & 79.7 & 80.6 \\
        7  & 78.0 & 79.6 & 81.6  \\
        \bottomrule 
    \end{tabular}
    \end{minipage}
\vspace{-10pt}
\end{table}

We also evaluated the performance with different \emph{Starts of the Second Training Stage (SSTS)} of the two-stage training scheme in our \interactionMechanism~in Table~\ref{tab-abltation_SSIT}, where $\text{SSTS}=N$ means that the first training stage is conducted over $N-1$ epochs. The results in the first row/block correspond to the standard one-stage training scheme in the IS interaction mechanism, which primarily focuses on correcting model self-errors. The results in the second block indicate that optimal performance is achieved when the first stage is completed in 4 epochs, further supporting the rationale for employing a two-stage learning approach in~\interactionMechanism.

%% file: sec/5_conclusion.tex
\vspace{-13pt}
\section{Conclusion} 
\label{sec:conclusion}
\vspace{-5pt}

In this paper, we introduce \interactiveMethodName, the first multi-scribble-guided deep learning framework for IOBE.
We additionally present two supporting components: \sytheticGenerationName, a tool for generating OB ground truth with explicit handling of self-occlusions from 3D scene mesh data; and \realBenchmarkName, a real-world benchmark for robust evaluation in OB research. 
Experimental results demonstrate that: (i) IOBE is highly effective for labeling OB GTs in monocular images; (ii) \interactiveMethodName~delivers promising performance by leveraging synthetic data without domain adaptation, with the potential for even better results under optimal conditions.
Despite these strengths, two main limitations remain: (i) The current human annotation experiment is still in the preliminary stage; and (ii) incomplete utilization of scribble interactions by the model, which can prevent their full integration into accurate predictions and cause interference among interactions in dense OB scenarios.
Building on the current framework, three key future directions include: (i) enhancing the model to improve IOBE performance; (ii) incorporating domain adaptation techniques for further optimization; (iii) conducting advanced OB GT annotation studies using IOBE.

%% file: bmvc_final.bbl
\begin{thebibliography}{78}
\providecommand{\natexlab}[1]{#1}
\providecommand{\url}[1]{\texttt{#1}}
\expandafter\ifx\csname urlstyle\endcsname\relax
  \providecommand{\doi}[1]{doi: #1}\else
  \providecommand{\doi}{doi: \begingroup \urlstyle{rm}\Url}\fi

\bibitem[Agustsson et~al.(2019)Agustsson, Uijlings, and Ferrari]{Agustsson_2019_CVPR}
Eirikur Agustsson, Jasper~RR Uijlings, and Vittorio Ferrari.
\newblock Interactive full image segmentation by considering all regions jointly.
\newblock In \emph{Proceedings of the IEEE/CVF Conference on Computer Vision and Pattern Recognition (CVPR)}, pages 11622--11631, 2019.

\bibitem[Andriluka et~al.(2020)Andriluka, Pellegrini, Popov, and Ferrari]{andriluka2020efficient}
Mykhaylo Andriluka, Stefano Pellegrini, Stefan Popov, and Vittorio Ferrari.
\newblock Efficient full image interactive segmentation by leveraging within-image appearance similarity.
\newblock \emph{arXiv preprint arXiv:2007.08173}, 2020.

\bibitem[Apostoloff and Fitzgibbon(2005)]{apostoloff2005learning}
Nicholas Apostoloff and Andrew Fitzgibbon.
\newblock Learning spatiotemporal t-junctions for occlusion detection.
\newblock In \emph{IEEE Computer Society Conference on Computer Vision and Pattern Recognition (CVPR)}, pages 553--559, 2005.

\bibitem[Black(1992)]{black1992combining}
Michael~J Black.
\newblock Combining intensity and motion for incremental segmentation and tracking over long image sequences.
\newblock In \emph{Proceedings of the European Conference on Computer Vision (ECCV)}, pages 485--493, 1992.

\bibitem[{Blender Online Community}(2018)]{blender_tool}
{Blender Online Community}.
\newblock \emph{Blender - a 3D modelling and rendering package}.
\newblock Blender Foundation, 2018.
\newblock URL \url{http://www.blender.org}.
\newblock Version 2.79b.

\bibitem[Castellani et~al.(2002)Castellani, Livatino, and Fisher]{castellani2002improving}
Umberto Castellani, Salvatore Livatino, and Robert~B Fisher.
\newblock Improving environment modelling by edge occlusion surface completion.
\newblock In \emph{Proceedings. First International Symposium on 3D Data Processing Visualization and Transmission}, pages 672--675, 2002.

\bibitem[Chen et~al.(2021)Chen, Zhao, Yu, Zhang, and Duan]{chen2021conditional}
Xi~Chen, Zhiyan Zhao, Feiwu Yu, Yilei Zhang, and Manni Duan.
\newblock Conditional diffusion for interactive segmentation.
\newblock In \emph{Proceedings of the IEEE/CVF International Conference on Computer Vision (ICCV)}, pages 7345--7354, 2021.

\bibitem[Chen et~al.(2022)Chen, Zhao, Zhang, Duan, Qi, and Zhao]{chen2022focalclick}
Xi~Chen, Zhiyan Zhao, Yilei Zhang, Manni Duan, Donglian Qi, and Hengshuang Zhao.
\newblock Focalclick: towards practical interactive image segmentation.
\newblock In \emph{Proceedings of the IEEE/CVF Conference on Computer Vision and Pattern Recognition (CVPR)}, pages 1300--1309, 2022.

\bibitem[Chen et~al.(2023)Chen, Cheung, Lim, and Zhao]{chen2023scribbleseg}
Xi~Chen, Yau Shing~Jonathan Cheung, Ser-Nam Lim, and Hengshuang Zhao.
\newblock Scribbleseg: Scribble-based interactive image segmentation.
\newblock \emph{arXiv preprint arXiv:2303.11320}, 2023.

\bibitem[Du et~al.(2023)Du, Yuan, Wang, and Wang]{du2023efficient}
Fei Du, Jianlong Yuan, Zhibin Wang, and Fan Wang.
\newblock Efficient mask correction for click-based interactive image segmentation.
\newblock In \emph{Proceedings of the IEEE/CVF Conference on Computer Vision and Pattern Recognition (CVPR)}, pages 22773--22782, 2023.

\bibitem[Dupont et~al.(2021)Dupont, Ouakrim, and Pham]{dupont2021ucp}
Camille Dupont, Yanis Ouakrim, and Quoc~Cuong Pham.
\newblock Ucp-net: unstructured contour points for instance segmentation.
\newblock In \emph{IEEE International Conference on Systems, Man, and Cybernetics (SMC)}, pages 3373--3379. IEEE, 2021.

\bibitem[Feldman and Weinshall(2008)]{feldman2008motion}
Doron Feldman and Daphna Weinshall.
\newblock Motion segmentation and depth ordering using an occlusion detector.
\newblock \emph{IEEE Transactions on Pattern Analysis and Machine Intelligence (TPAMI)}, 30\penalty0 (7):\penalty0 1171--1185, 2008.

\bibitem[Feng et~al.(2021)Feng, She, Zhu, Li, Zhang, Feng, Wang, Li, Kang, and Ming]{feng2021mt}
Panhe Feng, Qi~She, Lei Zhu, Jiaxin Li, Lin Zhang, Zijian Feng, Changhu Wang, Chunpeng Li, Xuejing Kang, and Anlong Ming.
\newblock Mt-orl: Multi-task occlusion relationship learning.
\newblock In \emph{Proceedings of the IEEE/CVF International Conference on Computer Vision (ICCV)}, pages 9364--9373, 2021.

\bibitem[Fu et~al.(2021)Fu, Jia, Gao, Gong, Zhao, Maybank, and Tao]{fu20213d}
Huan Fu, Rongfei Jia, Lin Gao, Mingming Gong, Binqiang Zhao, Steve Maybank, and Dacheng Tao.
\newblock 3d-future: 3d furniture shape with texture.
\newblock \emph{International Journal of Computer Vision (IJCV)}, 129:\penalty0 3313--3337, 2021.

\bibitem[Guo et~al.(2024)Guo, Shi, Li, Xiao, and Gao]{guo2024redir}
Qi~Guo, Hailong Shi, Huan Li, Jinsheng Xiao, and Xingyu Gao.
\newblock Redir: Refocus-free event-based de-occlusion image reconstruction.
\newblock In \emph{European Conference on Computer Vision}, pages 419--435. Springer, 2024.

\bibitem[Hambarde et~al.(2024)Hambarde, Wadhwa, Vipparthi, Murala, and Dhall]{hambarde2024occlusion}
Praful Hambarde, Gourav Wadhwa, Santosh~Kumar Vipparthi, Subrahmanyam Murala, and Abhinav Dhall.
\newblock Occlusion boundary prediction and transformer based depth-map refinement from single image.
\newblock \emph{ACM Transactions on Multimedia Computing, Communications and Applications}, 2024.

\bibitem[Hao et~al.(2021)Hao, Liu, Wu, Han, Chen, Chen, Chu, Tang, Yu, Chen, et~al.]{Hao_2021_ICCV_edgeflow}
Yuying Hao, Yi~Liu, Zewu Wu, Lin Han, Yizhou Chen, Guowei Chen, Lutao Chu, Shiyu Tang, Zhiliang Yu, Zeyu Chen, et~al.
\newblock Edgeflow: Achieving practical interactive segmentation with edge-guided flow.
\newblock In \emph{Proceedings of the IEEE/CVF International Conference on Computer Vision Workshops (ICCVW)}, pages 1551--1560, 2021.

\bibitem[Haouchine et~al.(2015)Haouchine, Dequidt, Berger, and Cotin]{haouchine2015monocular}
Nazim Haouchine, Jeremie Dequidt, Marie-Odile Berger, and Stephane Cotin.
\newblock Monocular 3d reconstruction and augmentation of elastic surfaces with self-occlusion handling.
\newblock \emph{IEEE Transactions on Visualization and Computer Graphics}, 21\penalty0 (12):\penalty0 1363--1376, 2015.

\bibitem[He and Yuille(2010)]{he2010occlusion}
Xuming He and Alan Yuille.
\newblock Occlusion boundary detection using pseudo-depth.
\newblock In \emph{Proceedings of the European Conference on Computer Vision (ECCV)}, pages 539--552, 2010.

\bibitem[Hoiem et~al.(2007)Hoiem, Stein, Efros, and Hebert]{hoiem2007recovering}
Derek Hoiem, Andrew~N Stein, Alexei~A Efros, and Martial Hebert.
\newblock Recovering occlusion boundaries from a single image.
\newblock In \emph{Proceedings of the IEEE/CVF International Conference on Computer Vision (ICCV)}, pages 1--8, 2007.

\bibitem[Hu et~al.(2019)Hu, Soltoggio, Lock, and Carter]{hu2019fully}
Yang Hu, Andrea Soltoggio, Russell Lock, and Steve Carter.
\newblock A fully convolutional two-stream fusion network for interactive image segmentation.
\newblock \emph{Neural Networks}, 109:\penalty0 31--42, 2019.

\bibitem[Huang et~al.(2024)Huang, Zou, Zhang, Bhatti, and Chen]{huang2024efficientmedical}
Mengxing Huang, Jie Zou, Yu~Zhang, Uzair~Aslam Bhatti, and Jing Chen.
\newblock Efficient click-based interactive segmentation for medical image with improved plain-vit.
\newblock \emph{IEEE Journal of Biomedical and Health Informatics}, pages 1--12, 2024.

\bibitem[Jacobson et~al.(2011)Jacobson, Freund, and Nguyen]{jacobson2011online}
Natan Jacobson, Yoav Freund, and Truong~Q Nguyen.
\newblock An online learning approach to occlusion boundary detection.
\newblock \emph{IEEE Transactions on Image Processing (TIP)}, 21\penalty0 (1):\penalty0 252--261, 2011.

\bibitem[Jang and Kim(2019)]{jang2019interactive}
Won-Dong Jang and Chang-Su Kim.
\newblock Interactive image segmentation via backpropagating refinement scheme.
\newblock In \emph{Proceedings of the IEEE/CVF Conference on Computer Vision and Pattern Recognition (CVPR)}, pages 5297--5306, 2019.

\bibitem[Jia et~al.(2012)Jia, Gallagher, Chang, and Chen]{jia2012learning}
Zhaoyin Jia, Andrew Gallagher, Yao-Jen Chang, and Tsuhan Chen.
\newblock A learning-based framework for depth ordering.
\newblock In \emph{Proceedings of the IEEE/CVF Conference on Computer Vision and Pattern Recognition (CVPR)}, pages 294--301, 2012.

\bibitem[Karmann and Urfalioglu(2025)]{M2N22025}
Markus Karmann and Onay Urfalioglu.
\newblock Repurposing stable diffusion attention for training-free unsupervised interactive segmentation.
\newblock In \emph{Proceedings of the IEEE/CVF Conference on Computer Vision and Pattern Recognition (CVPR)}, June 2025.

\bibitem[Karsch et~al.(2013)Karsch, Liao, Rock, Barron, and Hoiem]{Karsch_2013_CVPR}
Kevin Karsch, Zicheng Liao, Jason Rock, Jonathan~T Barron, and Derek Hoiem.
\newblock Boundary cues for 3d object shape recovery.
\newblock In \emph{Proceedings of the IEEE Conference on Computer Vision and Pattern Recognition (CVPR)}, pages 2163--2170, 2013.

\bibitem[Keller et~al.(2007)Keller, Knothe, and Vetter]{keller20073d}
Michael Keller, Reinhard Knothe, and Thomas Vetter.
\newblock 3d reconstruction of human faces from occluding contours.
\newblock In \emph{Computer Vision/Computer Graphics Collaboration Techniques}, pages 261--273, 2007.

\bibitem[Khirodkar et~al.(2022)Khirodkar, Tripathi, and Kitani]{khirodkar2022occluded}
Rawal Khirodkar, Shashank Tripathi, and Kris Kitani.
\newblock Occluded human mesh recovery.
\newblock In \emph{Proceedings of the IEEE/CVF Conference on Computer Vision and Pattern Recognition (CVPR)}, pages 1715--1725, 2022.

\bibitem[Koch et~al.(2018)Koch, Liebel, Fraundorfer, and Korner]{koch2018evaluation}
Tobias Koch, Lukas Liebel, Friedrich Fraundorfer, and Marco Korner.
\newblock Evaluation of cnn-based single-image depth estimation methods.
\newblock In \emph{Proceedings of the European Conference on Computer Vision Workshops (ECCVW)}, pages 0--0, 2018.

\bibitem[Le et~al.(2018)Le, Mai, Price, Cohen, Jin, and Liu]{le2018interactive}
Hoang Le, Long Mai, Brian Price, Scott Cohen, Hailin Jin, and Feng Liu.
\newblock Interactive boundary prediction for object selection.
\newblock In \emph{Proceedings of the European Conference on Computer Vision (ECCV)}, pages 18--33, 2018.

\bibitem[Lee et~al.(2024)Lee, Lee, and Kim]{lee2024mfp}
Chaewon Lee, Seon-Ho Lee, and Chang-Su Kim.
\newblock Mfp: Making full use of probability maps for interactive image segmentation.
\newblock In \emph{2024 IEEE/CVF Conference on Computer Vision and Pattern Recognition (CVPR)}, pages 4051--4059. IEEE, 2024.

\bibitem[Lee et~al.(2022)Lee, Park, Song, Ryu, Kim, Kim, Pereira, and Yoo]{lee2022interactive}
Chunggi Lee, Seonwook Park, Heon Song, Jeongun Ryu, Sanghoon Kim, Haejoon Kim, S{\'e}rgio Pereira, and Donggeun Yoo.
\newblock Interactive multi-class tiny-object detection.
\newblock In \emph{Proceedings of the IEEE/CVF Conference on Computer Vision and Pattern Recognition (CVPR)}, pages 14136--14145, 2022.

\bibitem[Li et~al.(2019)Li, Gao, and Wu]{li2019high}
Jianwei Li, Wei Gao, and Yihong Wu.
\newblock High-quality 3d reconstruction with depth super-resolution and completion.
\newblock \emph{IEEE Access}, 7:\penalty0 19370--19381, 2019.

\bibitem[Li and Chen(2022)]{li2022efficient}
Ruizhe Li and Xin Chen.
\newblock An efficient interactive multi-label segmentation tool for 2d and 3d medical images using fully connected conditional random field.
\newblock \emph{Computer Methods and Programs in Biomedicine}, 213:\penalty0 106534, 2022.

\bibitem[Li et~al.(2018)Li, Saeedi, McCormac, Clark, Tzoumanikas, Ye, Huang, Tang, and Leutenegger]{li2018interiornet}
Wenbin Li, Sajad Saeedi, John McCormac, Ronald Clark, Dimos Tzoumanikas, Qing Ye, Yuzhong Huang, Rui Tang, and Stefan Leutenegger.
\newblock Interiornet: Mega-scale multi-sensor photo-realistic indoor scenes dataset.
\newblock \emph{arXiv preprint arXiv:1809.00716}, 2018.

\bibitem[Li et~al.(2023)Li, Long, Wang, Cao, Wang, Luo, and Xiao]{li2023neto}
Zongcheng Li, Xiaoxiao Long, Yusen Wang, Tuo Cao, Wenping Wang, Fei Luo, and Chunxia Xiao.
\newblock Neto: neural reconstruction of transparent objects with self-occlusion aware refraction-tracing.
\newblock In \emph{Proceedings of the IEEE/CVF International Conference on Computer Vision (ICCV)}, pages 18547--18557, 2023.

\bibitem[Liew et~al.(2021)Liew, Cohen, Price, Mai, and Feng]{liew2021deepthin}
Jun~Hao Liew, Scott Cohen, Brian Price, Long Mai, and Jiashi Feng.
\newblock Deep interactive thin object selection.
\newblock In \emph{Proceedings of the IEEE/CVF Winter Conference on Applications of Computer Vision (WACV)}, pages 305--314, 2021.

\bibitem[Lin et~al.(2025)Lin, Chen, Yang, Roitberg, Li, Li, and Li]{lin2025adapclick}
Jiacheng Lin, Jiajun Chen, Kailun Yang, Alina Roitberg, Siyu Li, Zhiyong Li, and Shutao Li.
\newblock Adaptiveclick: Click-aware transformer with adaptive focal loss for interactive image segmentation.
\newblock \emph{IEEE Transactions on Neural Networks and Learning Systems}, 36\penalty0 (3):\penalty0 5759--5773, 2025.

\bibitem[Lin et~al.(2020)Lin, Zhang, Chen, Cheng, and Lu]{lin2020fcanet}
Zheng Lin, Zhao Zhang, Lin-Zhuo Chen, Ming-Ming Cheng, and Shao-Ping Lu.
\newblock Interactive image segmentation with first click attention.
\newblock In \emph{Proceedings of the IEEE/CVF Conference on Computer Vision and Pattern Recognition (CVPR)}, pages 13339--13348, 2020.

\bibitem[Liu et~al.(2023)Liu, Xu, Bertasius, and Niethammer]{liu2023simpleclick}
Qin Liu, Zhenlin Xu, Gedas Bertasius, and Marc Niethammer.
\newblock Simpleclick: Interactive image segmentation with simple vision transformers.
\newblock In \emph{Proceedings of the IEEE/CVF International Conference on Computer Vision (ICCV)}, pages 22290--22300, 2023.

\bibitem[Liu et~al.(2021)Liu, Lin, Cao, Hu, Wei, Zhang, Lin, and Guo]{liu2021swin}
Ze~Liu, Yutong Lin, Yue Cao, Han Hu, Yixuan Wei, Zheng Zhang, Stephen Lin, and Baining Guo.
\newblock Swin transformer: Hierarchical vision transformer using shifted windows.
\newblock In \emph{Proceedings of the IEEE/CVF International Conference on Computer Vision (ICCV)}, pages 10012--10022, 2021.

\bibitem[Lu et~al.(2023)Lu, Kuen, Tiancheng, Jiuxiang, Weidong, Jiaya, Zhe, and Ming-Hsuan]{qi2022fine}
Qi~Lu, Jason Kuen, Shen Tiancheng, Gu~Jiuxiang, Guo Weidong, Jia Jiaya, Lin Zhe, and Yang Ming-Hsuan.
\newblock High-quality entity segmentation.
\newblock In \emph{Proceedings of the IEEE/CVF International Conference on Computer Vision (ICCV)}, pages 4047--4056, 2023.

\bibitem[Lu et~al.(2019)Lu, Xue, Zhou, Ming, and Zhou]{lu2019occlusion}
Rui Lu, Feng Xue, Menghan Zhou, Anlong Ming, and Yu~Zhou.
\newblock Occlusion-shared and feature-separated network for occlusion relationship reasoning.
\newblock In \emph{Proceedings of the IEEE/CVF International Conference on Computer Vision (ICCV)}, pages 10343--10352, 2019.

\bibitem[Luo et~al.(2021)Luo, Wang, Song, Zhang, Aertsen, Deprest, Ourselin, Vercauteren, and Zhang]{luo2021mideepseg}
Xiangde Luo, Guotai Wang, Tao Song, Jingyang Zhang, Michael Aertsen, Jan Deprest, Sebastien Ourselin, Tom Vercauteren, and Shaoting Zhang.
\newblock Mideepseg: Minimally interactive segmentation of unseen objects from medical images using deep learning.
\newblock \emph{Medical Image Analysis}, 72:\penalty0 102102, 2021.

\bibitem[Majumder et~al.(2020)Majumder, Rai, Khurana, and Yao]{majumder2020two}
Soumajit Majumder, Abhinav Rai, Ansh Khurana, and Angela Yao.
\newblock Two-in-one refinement for interactive segmentation.
\newblock In \emph{British Machine Vision Conference (BMVC)}, page~2, 2020.

\bibitem[Maninis et~al.(2018)Maninis, Caelles, Pont-Tuset, and Van~Gool]{maninis2018deep}
Kevis-Kokitsi Maninis, Sergi Caelles, Jordi Pont-Tuset, and Luc Van~Gool.
\newblock Deep extreme cut: From extreme points to object segmentation.
\newblock In \emph{Proceedings of the IEEE Conference on Computer Vision and Pattern Recognition (CVPR)}, pages 616--625, 2018.

\bibitem[Ming et~al.(2015)Ming, Wu, Ma, Sun, and Zhou]{ming2015monocular}
Anlong Ming, Tianfu Wu, Jianxiang Ma, Fang Sun, and Yu~Zhou.
\newblock Monocular depth-ordering reasoning with occlusion edge detection and couple layers inference.
\newblock \emph{IEEE Intelligent Systems}, 31\penalty0 (2):\penalty0 54--65, 2015.

\bibitem[Myers-Dean et~al.(2024)Myers-Dean, Fan, Price, Chan, and Gurari]{Myers-Dean_2024_WACV}
Josh Myers-Dean, Yifei Fan, Brian Price, Wilson Chan, and Danna Gurari.
\newblock Interactive segmentation for diverse gesture types without context.
\newblock In \emph{Proceedings of the IEEE/CVF Winter Conference on Applications of Computer Vision (WACV)}, pages 7198--7208, 2024.

\bibitem[Nieuwenhuis and Cremers(2012)]{nieuwenhuis2012spatially}
Claudia Nieuwenhuis and Daniel Cremers.
\newblock Spatially varying color distributions for interactive multilabel segmentation.
\newblock \emph{IEEE Transactions on Pattern Analysis and Machine Intelligence (TPAMI)}, 35\penalty0 (5):\penalty0 1234--1247, 2012.

\bibitem[Nieuwenhuis et~al.(2014)Nieuwenhuis, Hawe, Kleinsteuber, and Cremers]{nieuwenhuis2014co}
Claudia Nieuwenhuis, Simon Hawe, Martin Kleinsteuber, and Daniel Cremers.
\newblock Co-sparse textural similarity for interactive segmentation.
\newblock In \emph{Proceedings of the European Conference on Computer Vision (ECCV)}, pages 285--301, 2014.

\bibitem[Popenova et~al.(2023)Popenova, Galeev, Vorontsova, and Konushin]{galeev2023contour}
Polina Popenova, Danil Galeev, Anna Vorontsova, and Anton Konushin.
\newblock Contour-based interactive segmentation.
\newblock In \emph{Proceedings of the Thirty-Second International Joint Conference on Artificial Intelligence}, pages 1322--1330, 2023.

\bibitem[Pu et~al.(2021)Pu, Huang, Guan, and Ling]{pu2021rindnet}
Mengyang Pu, Yaping Huang, Qingji Guan, and Haibin Ling.
\newblock Rindnet: Edge detection for discontinuity in reflectance, illumination, normal and depth.
\newblock In \emph{Proceedings of the IEEE/CVF International Conference on Computer Vision (ICCV)}, pages 6879--6888, 2021.

\bibitem[Qiu et~al.(2020)Qiu, Xiao, Wang, and Marlet]{qiu2020pixel}
Xuchong Qiu, Yang Xiao, Chaohui Wang, and Renaud Marlet.
\newblock Pixel-pair occlusion relationship map (p2orm): formulation, inference and application.
\newblock In \emph{Proceedings of the European Conference on Computer Vision (ECCV)}, pages 690--708, 2020.

\bibitem[Ramamonjisoa and Lepetit(2019)]{ramamonjisoa2019sharpnet}
Michael Ramamonjisoa and Vincent Lepetit.
\newblock Sharpnet: Fast and accurate recovery of occluding contours in monocular depth estimation.
\newblock In \emph{Proceedings of the IEEE/CVF International Conference on Computer Vision Workshops (ICCVW)}, pages 0--0, 2019.

\bibitem[Ramamonjisoa et~al.(2020)Ramamonjisoa, Du, and Lepetit]{Ramamonjisoa_2020_CVPR}
Michael Ramamonjisoa, Yuming Du, and Vincent Lepetit.
\newblock Predicting sharp and accurate occlusion boundaries in monocular depth estimation using displacement fields.
\newblock In \emph{Proceedings of the IEEE/CVF Conference on Computer Vision and Pattern Recognition (CVPR)}, pages 14648--14657, 2020.

\bibitem[Raskar et~al.(2004)Raskar, Tan, Feris, Yu, and Turk]{raskar2004non}
Ramesh Raskar, Kar-Han Tan, Rogerio Feris, Jingyi Yu, and Matthew Turk.
\newblock Non-photorealistic camera: depth edge detection and stylized rendering using multi-flash imaging.
\newblock \emph{ACM Transactions on Graphics (TOG)}, 23\penalty0 (3):\penalty0 679--688, 2004.

\bibitem[Ren et~al.(2006)Ren, Fowlkes, and Malik]{ren2006figure}
Xiaofeng Ren, Charless~C Fowlkes, and Jitendra Malik.
\newblock Figure/ground assignment in natural images.
\newblock In \emph{Proceedings of the European Conference on Computer Vision (ECCV)}, pages 614--627, 2006.

\bibitem[Santner et~al.(2011)Santner, Pock, and Bischof]{santner2011interactive}
Jakob Santner, Thomas Pock, and Horst Bischof.
\newblock Interactive multi-label segmentation.
\newblock In \emph{Asian Conference on Computer Vision (ACCV)}, pages 397--410, 2011.

\bibitem[Sargin et~al.(2009)Sargin, Bertelli, Manjunath, and Rose]{sargin2009probabilistic}
Mehmet~Emre Sargin, Luca Bertelli, Bangalore~S Manjunath, and Kenneth Rose.
\newblock Probabilistic occlusion boundary detection on spatio-temporal lattices.
\newblock In \emph{Proceedings of the IEEE/CVF International Conference on Computer Vision (ICCV)}, pages 560--567, 2009.

\bibitem[Sofiiuk et~al.(2022)Sofiiuk, Petrov, and Konushin]{sofiiuk2022reviving}
Konstantin Sofiiuk, Ilya~A Petrov, and Anton Konushin.
\newblock Reviving iterative training with mask guidance for interactive segmentation.
\newblock In \emph{IEEE International Conference on Image Processing (ICIP)}, pages 3141--3145, 2022.

\bibitem[Stein and Hebert(2009)]{stein2009occlusion}
Andrew~N Stein and Martial Hebert.
\newblock Occlusion boundaries from motion: Low-level detection and mid-level reasoning.
\newblock \emph{International Journal of Computer Vision (IJCV)}, 82:\penalty0 325--357, 2009.

\bibitem[Sundberg et~al.(2011)Sundberg, Brox, Maire, Arbel{\'a}ez, and Malik]{sundberg2011occlusion}
Patrik Sundberg, Thomas Brox, Michael Maire, Pablo Arbel{\'a}ez, and Jitendra Malik.
\newblock Occlusion boundary detection and figure/ground assignment from optical flow.
\newblock In \emph{IEEE Computer Society Conference on Computer Vision and Pattern Recognition (CVPR)}, pages 2233--2240, 2011.

\bibitem[Teo et~al.(2015)Teo, Fermuller, and Aloimonos]{teo2015fast}
Ching Teo, Cornelia Fermuller, and Yiannis Aloimonos.
\newblock Fast 2d border ownership assignment.
\newblock In \emph{Proceedings of the IEEE Conference on Computer Vision and Pattern Recognition (CVPR)}, pages 5117--5125, 2015.

\bibitem[Vasiljevic et~al.(2019)Vasiljevic, Kolkin, Zhang, Luo, Wang, Dai, Daniele, Mostajabi, Basart, Walter, et~al.]{vasiljevic2019diode}
Igor Vasiljevic, Nick Kolkin, Shanyi Zhang, Ruotian Luo, Haochen Wang, Falcon~Z Dai, Andrea~F Daniele, Mohammadreza Mostajabi, Steven Basart, Matthew~R Walter, et~al.
\newblock Diode: A dense indoor and outdoor depth dataset.
\newblock \emph{arXiv preprint arXiv:1908.00463}, 2019.

\bibitem[Vezhnevets and Konouchine(2005)]{vezhnevets2005growcut}
Vladimir Vezhnevets and Vadim Konouchine.
\newblock Growcut: Interactive multi-label nd image segmentation by cellular automata.
\newblock In \emph{Graphicon}, pages 150--156, 2005.

\bibitem[Wang et~al.(2020)Wang, Fu, Tao, and Black]{wang2020occlusion}
Chaohui Wang, Huan Fu, Dacheng Tao, and Michael~J Black.
\newblock Occlusion boundary: A formal definition \& its detection via deep exploration of context.
\newblock \emph{IEEE Transactions on Pattern Analysis and Machine Intelligence (TPAMI)}, 44\penalty0 (5):\penalty0 2641--2656, 2020.

\bibitem[Wang et~al.(2018)Wang, Li, Zuluaga, Pratt, Patel, Aertsen, Doel, David, Deprest, Ourselin, et~al.]{wang2018interactive}
Guotai Wang, Wenqi Li, Maria~A Zuluaga, Rosalind Pratt, Premal~A Patel, Michael Aertsen, Tom Doel, Anna~L David, Jan Deprest, S{\'e}bastien Ourselin, et~al.
\newblock Interactive medical image segmentation using deep learning with image-specific fine tuning.
\newblock \emph{IEEE Transactions on Medical Imaging}, 37\penalty0 (7):\penalty0 1562--1573, 2018.

\bibitem[Wang et~al.(2019)Wang, Wang, Li, and Liang]{wang2019doobnet}
Guoxia Wang, Xiaochuan Wang, Frederick~WB Li, and Xiaohui Liang.
\newblock Doobnet: Deep object occlusion boundary detection from an image.
\newblock In \emph{Asian Conference on Computer Vision (ACCV)}, pages 686--702, 2019.

\bibitem[Wang and Yuille(2016)]{wang2016doc}
Peng Wang and Alan Yuille.
\newblock Doc: Deep occlusion estimation from a single image.
\newblock In \emph{Proceedings of the European Conference on Computer Vision (ECCV)}, pages 545--561, 2016.

\bibitem[Wu et~al.(2025)Wu, Wang, Yang, Liu, Zeng, Ye, and Li]{wu2025occfree_cvpr}
You Wu, Xucheng Wang, Xiangyang Yang, Mengyuan Liu, Dan Zeng, Hengzhou Ye, and Shuiwang Li.
\newblock Learning occlusion-robust vision transformers for real-time uav tracking.
\newblock \emph{arXiv preprint arXiv:2504.09228}, 2025.

\bibitem[Xu et~al.(2025)Xu, Wang, and Wang]{xu2025modot}
Lintao Xu, Yinghao Wang, and Chaohui Wang.
\newblock Occlusion boundary and depth: Mutual enhancement via multi-task learning.
\newblock \emph{arXiv preprint arXiv:2505.21231}, 2025.

\bibitem[Xu et~al.(2016)Xu, Price, Cohen, Yang, and Huang]{xu2016deep}
Ning Xu, Brian Price, Scott Cohen, Jimei Yang, and Thomas~S Huang.
\newblock Deep interactive object selection.
\newblock In \emph{Proceedings of the IEEE Conference on Computer Vision and Pattern Recognition (CVPR)}, pages 373--381, 2016.

\bibitem[Yang et~al.(2023)Yang, Kong, Min, Wee, Jang, Cha, and Kang]{yang2023sefd}
ChangHee Yang, Kyeongbo Kong, SungJun Min, Dongyoon Wee, Ho-Deok Jang, Geonho Cha, and SukJu Kang.
\newblock Sefd: learning to distill complex pose and occlusion.
\newblock In \emph{Proceedings of the IEEE/CVF International Conference on Computer Vision (ICCV)}, pages 14941--14952, 2023.

\bibitem[Zhang et~al.(2020)Zhang, Liew, Wei, Wei, and Zhao]{zhang2020interactive}
Shiyin Zhang, Jun~Hao Liew, Yunchao Wei, Shikui Wei, and Yao Zhao.
\newblock Interactive object segmentation with inside-outside guidance.
\newblock In \emph{Proceedings of the IEEE/CVF Conference on Computer Vision and Pattern Recognition (CVPR)}, pages 12234--12244, 2020.

\bibitem[Zhao et~al.(2024)Zhao, Li, Cheng, Qiao, Zheng, Ji, Liu, Yuan, and Chen]{Zhao2024graco}
Yian Zhao, Kehan Li, Zesen Cheng, Pengchong Qiao, Xiawu Zheng, Rongrong Ji, Chang Liu, Li~Yuan, and Jie Chen.
\newblock Graco: Granularity-controllable interactive segmentation.
\newblock In \emph{Proceedings of the IEEE/CVF Conference on Computer Vision and Pattern Recognition (CVPR)}, pages 3501--3510, 2024.

\bibitem[Zhou et~al.(2023)Zhou, Wang, Zhao, Li, Huang, Meng, and Zheng]{zhou2023interactive}
Minghao Zhou, Hong Wang, Qian Zhao, Yuexiang Li, Yawen Huang, Deyu Meng, and Yefeng Zheng.
\newblock Interactive segmentation as gaussion process classification.
\newblock In \emph{Proceedings of the IEEE/CVF Conference on Computer Vision and Pattern Recognition (CVPR)}, pages 19488--19497, 2023.

\bibitem[Zhu et~al.(2017)Zhu, Wang, and Yu]{zhu2017occlusion}
Hao Zhu, Qing Wang, and Jingyi Yu.
\newblock Occlusion-model guided antiocclusion depth estimation in light field.
\newblock \emph{IEEE Journal of Selected Topics in Signal Processing}, 11\penalty0 (7):\penalty0 965--978, 2017.

\end{thebibliography}
